%% file: main.tex
\documentclass[10pt,twocolumn,letterpaper]{article}
\usepackage[pagenumbers]{cvpr}

\usepackage{amsmath,amssymb,graphicx,booktabs,array}
\usepackage{microtype,placeins,url,needspace}
\definecolor{cvprblue}{rgb}{0.21,0.49,0.74}
\usepackage[breaklinks,colorlinks,allcolors=cvprblue]{hyperref}
\hypersetup{
    pdftitle={Bootstrap Once, Then Update Conservatively: Simplifying State-based Online 3D Reconstruction},
    pdfauthor={Lei Zhou, Haoyu Wu, Akshat Dave, Dimitris Samaras}
}

\title{Bootstrap Once, Then Update Conservatively:\\
 Simplifying State-based Online 3D Reconstruction}

\author{%
  Lei Zhou\textsuperscript{1}\thanks{\href{mailto:lezzhou@cs.stonybrook.edu}{\texttt{lezzhou@cs.stonybrook.edu}}}\qquad
  Haoyu Wu\textsuperscript{2}\qquad
  Akshat Dave\textsuperscript{1}\qquad
  Dimitris Samaras\textsuperscript{1}\\[3pt]
  \textsuperscript{1}Stony Brook University\qquad \textsuperscript{2}Amazon}

\makeatletter
\g@addto@macro\@maketitle{\input{figures/teaser}\par\vspace{0.5\baselineskip}}
\makeatother

\begin{document}
\maketitle

\begin{abstract}
State-based Online 3D reconstruction maintains a persistent memory that
integrates multi-view information from past observations to support predictions for the current frame.
Vanilla pretrained backbones apply their predicted state updates using a full weight.
Recent methods improve performance by adapting the update weight to the current observation and state at test time.
However, they make two changes at once: they reduce the average update weight and vary that weight across frames or memory components. 
Thus, we ask:
how much does this adaptive modulation help?
To answer this question, we introduce Bootstrap Once, Then Update Conservatively (BootCon). 
BootCon commits the first update in full to initialize a scene-conditioned memory, then applies a small, constant weight to every subsequent update. 
We find that calibrating this constant weight to the mean of each prior adaptive method's update weight, almost matches its improvement.
We then propose a simple heuristic that directly sets the update weight to the ratio of the training context length to the inference context length. In this
setting, BootCon outperforms the adaptive methods on all tasks per primary metrics, 
establishing it as an effective
practical approach for online 3D reconstruction, without any adaptive modulation.
\end{abstract}

\section{Introduction}
\label{sec:introduction}
Online 3D reconstruction estimates camera motion and scene geometry from
a stream of images. Persistent memory allows a model to use information
from earlier views when processing each new frame. CUT3R represents this
memory with scene tokens~\citep{wang2025cut3r}, while streaming ZipMap
uses the fast weights of test-time-training MLPs~\citep{jin2026zipmap}.
Memory updates must incorporate new observations while preserving useful
information from earlier frames. Choosing how strongly to update the
memory is therefore crucial to accurate and consistent reconstruction
over long sequences.

Vanilla pretrained backbones apply the full predicted state updates
using an update weight of one. 
Recent methods regulate memory through adaptive update weights.
TTT3R~\citep{chen2026ttt3r} uses state--observation attention to determine
token-wise update weights, while FILT3R~\citep{jin2026filt3r} adjusts them
through drift-dependent filtering. These methods improve substantially
over their vanilla backbones. However, they make two changes at once:
they reduce the \emph{average update weight} and vary that weight
across frames or memory components. Comparing only against full updates
does not distinguish the benefit of conservative updating from that of
adaptive modulation. We therefore ask:
\emph{How much does this adaptive modulation help?}

To answer this question, we introduce \emph{Bootstrap Once, Then Update Conservatively
(BootCon)}. BootCon commits the first update in full to initialize a
scene-conditioned memory, then applies a small, constant weight $c$
to every subsequent update. 
It applies to base models, such as CUT3R and streaming ZipMap, without retraining or
changing the backbone's update proposal. 

In a control experiment, we first calibrate $c$ to each adaptive reference's mean
update weight after the first frame.
Across six benchmarks covering camera-pose estimation, video-depth
prediction, and 3D reconstruction, mean-matched BootCon improves baseline performance almost as much as or even more than
($94\%$--$124\%$) the adaptive methods
(Table~\ref{tab:bootcon_mean_matched}). 
This result shifts the focus from designing
more elaborate modulation to understanding why conservative updating
is already so effective.

We investigate this behavior through \emph{stability and plasticity
probing}. Stability asks whether memory still reproduces its original
first-window predictions as subsequent context grows. Plasticity asks how
much updating on a fixed last window improves its reconstruction as
preceding context grows. 
Mean-matched BootCon broadly follows each adaptive method's
stability and plasticity patterns (Figure~\ref{fig:diagnostics}),
suggesting that similar memory dynamics can be realized without adaptive modulation. The probes are consistent with the
intuition that conservative writes limit accumulated update errors and
interference while retaining capability of learning from incoming observations.

Reproducing an adaptive method
does not imply that its average weight is well chosen. 
We then propose a simple heuristic that directly sets $c$
to the ratio of the training context length to the inference context length.
This spreads roughly a training-length total update
budget over the intended inference horizon.
BootCon with this ratio reduces the absolute trajectory error on the TUM dataset by $71\%$
relative to vanilla CUT3R and the scale-aligned depth error on the Bonn dataset by $43\%$
relative to vanilla ZipMap. With benchmark-specific design horizons,
ZipMap-based BootCon outperforms all nine comparison methods on each
benchmark's primary metric. On 7-Scenes it achieves the lowest mean
accuracy and completeness errors (Table~\ref{tab:bootcon_context_comparison}).
Thus, BootCon is an effective practical approach for
online 3D reconstruction, obviating the need for adaptive modulation.

Our contributions are:
\begin{itemize}
    \item We introduce BootCon, a simple approach for setting update weight in state-based 3D reconstruction models. 
    We find matching BootCon's constant update weight with the adaptive methods' average update weight preserves most of their performance gains.
    \item We use fixed-window stability--plasticity probes to show that
    mean-matched BootCon broadly follows adaptive methods'
    patterns, suggesting similar memory dynamics.
    \item We propose to set BootCon's constant update weight to the ratio of the training context length to the inference context length. In this setting,
    ZipMap-based BootCon outperforms the adaptive methods on all benchmarks.
\end{itemize}

\section{Related Work}
\label{sec:related_work}
\paragraph{Online 3D reconstruction.}
Recent feed-forward models~\citep{wang2025vggt,wang2025pi3,keetha2026mapanything,lin2025da3,wang2026vggtomega} infer camera parameters and dense geometry directly from images.
Long-sequence extensions combine local submaps with geometric alignment and loop-aware optimization~\citep{maggio2025vggtslam,deng2025vggtlong,lee2025swiftvggt,maggio2026vggtslam2}.
Further developments address motion-aware submap construction and spatially varying inconsistencies~\citep{xiong2026vggtmotion,zhang2026talo}, strengthen registration and back-end consistency~\citep{mandal2026vggtslampp,zhang2026vggtalign,choi2026vggtbridge,wang2026amb3rslam,li2026grfrecon}, or learn local-to-world registration~\citep{liu2025slam3r}.
Although their causal protocols differ, these methods explicitly reconcile local reconstructions.
Persistent-state approaches instead encode history in scene tokens, fast weights, or hybrid memories~\citep{wang2025cut3r,jin2026zipmap,liu2026mem3r}.
We study causal updates to CUT3R's scene tokens and streaming ZipMap's fast weights, rather than submap stitching or back-end optimization.

\paragraph{Adaptive memory updates for 3D reconstruction.}
Adaptive updates draw on state--observation alignment~\citep{chen2026ttt3r}, filtering~\citep{haarnoja2016backpropkf,jin2026filt3r}, temporal and spatial reliability~\citep{zheng2026ttsa3r}, and selective patch, frame, or motion-aware control~\citep{dong2026memix,ren2026afg,xu2026pas3r}.
Conservative writing is already an explicit concern.
FILT3R supports its adaptive-filtering design with a tuned constant-update baseline that degrades relative rotation and metric depth, but this comparison does not match the average update weight.
Our mean-matched control challenges the necessity of this trade-off: on TUM, BootCon retains over $94\%$ of FILT3R's ATE improvement while slightly improving both relative-motion metrics (Tables~\ref{tab:bootcon_mean_matched} and~\ref{tab:bootcon_rpe_matched}).
Thus, the reported failure of constant updating is not intrinsic to removing adaptivity.
ReCal3R combines a conservative base rate with reliability-dependent modulation~\citep{li2026recal3r}; its ablations favor small nonzero fallback rates over freezing, but retain adaptivity rather than testing mean-matched constants.
BootCon separates these effects by preserving the full first update and removing subsequent modulation while matching its mean weight.
We further select an effective constant directly from the training-to-inference context ratio.

\paragraph{Fast weights and test-time training.}
Fast-weight methods connect linear attention to online learning and treat a trainable model as recurrent memory updated through a self-supervised objective~\citep{schlag2021fastweights,sun2025ttt}.
Subsequent work improves update efficiency and learning objectives~\citep{zhang2026lact,feng2026inplace}, and extends fast-weight memory to robot policies~\citep{jiang2026robottt}.
Scene-specific neural mapping~\citep{sucar2021imap} and long-context Gaussian reconstruction~\citep{wang2026tttlrm} also adapt MLP parameters or fast weights during inference.
The latter refines past predictions from posed observations, unlike emitting fixed pose and point predictions from an unposed stream.
These works develop memory representations and update rules; we keep the backbone's proposal intact and study how strongly it should be committed.
A constant update weight does not make that proposal observation-independent.

\section{State-based Online 3D Reconstruction}
\label{sec:prelim}
Given an image stream $\{I_1,I_2,\ldots\}$, a stateful reconstruction model
predicts camera poses and scene geometry using the current observation and
a memory carried from previous frames. 
In this paper, we consider two recurrent backbones: 
CUT3R~\citep{wang2025cut3r}, with scene tokens as state, 
and ZipMap~\citep{jin2026zipmap}, with fast weights as state.
Below, we unify how updates are \emph{committed} under these two backbones.

\subsection{Recurrent Backbones}
\label{sec:recurrent_backbones}
Let $S_{t-1}$ denote the persistent state, i.e., memory, before frame $t$, 
which is initialized from the pretrained state $S_0$.
The state transition from $S_{t-1}$ to $S_t$ is given by:
\begin{equation}
    S_t = \Pi_t\!\left(S_{t-1}+\Delta_t\right),
    \label{eq:vanilla_commit}
\end{equation}
where $\Pi_t$ denotes post-update projection onto the backbone's state
constraint manifold, up to numerical stabilization. For unconstrained
states, $\Pi_t$ is the identity.
The update $\Delta_t$ is predicted by the backbone directly or indirectly.
\\
\textbf{Scene Tokens as State.} CUT3R~\citep{wang2025cut3r} represents scene memory as
$S_t\in\mathbb R^{N\times d}$, with $N$ latent tokens of dimension $d$.
Its recurrent decoder produces a candidate state
$\widetilde S_t=F(S_{t-1},I_t)$, where $F$ denotes the frozen
decoder. 
Thus, under the formulation of Equation~\ref{eq:vanilla_commit}, 
$\Delta_t=\widetilde S_t-S_{t-1}$ and
$\Pi_t=\operatorname{Identity}$. 
The current frame's predictions are decoded before the scene state update
is committed. 
\\
\textbf{Fast Weights as State.}
ZipMap~\citep{jin2026zipmap} instead stores memory in the fast
weights of test-time-training MLPs. Let $W_t$ denote one such weight matrix in the MLP. 
The update is $W_t=\mathcal N_t(W_{t-1} + G_t)$ where $G_t=\mathcal G_{\mathtt{Muon}}(W_{t-1},I_t)$.
Here $\mathcal G_{\mathtt{Muon}}$ computes the Hebbian gradient with Muon optimizer;
$I_t$ denotes the current observation's tokens at time $t$;
$\mathcal N_t$ restores the vector norms along the pre-defined axis.
The unified terms are therefore $S_t=W_t$,
$\Delta_t=G_t$, and $\Pi_t=\mathcal N_t$.
Since we focus on online reconstruction, we use 
ZipMap's official streaming checkpoint for our model initialization. 
Appendix~\ref{app:zipmap_streaming} details the per-frame write, normalization,
and prediction order.

\subsection{Adaptive Variants}
\label{sec:adaptive_variants}
Adaptive methods introduce a \emph{coefficient} $a_t$ that modulates
the update weight:
\begin{equation}
    S_t=\Pi_t\!\left(S_{t-1}+a_t\odot\Delta_t\right).
    \label{eq:adaptive}
\end{equation}
The coefficients are broadcast over the dimensions they control.
For scene token $i$ in CUT3R, this gives
$S_{t,i}=(1-a_{t,i})S_{t-1,i}+a_{t,i}\widetilde S_{t,i}$;
for a fast-weight matrix in ZipMap, it gives
$W_t=\mathcal N_t(W_{t-1} + a_t G_t)$.
Unit coefficients, $a_t=1$, recover the vanilla updates.
Coefficients describe how strongly a proposal is committed.
Next, we introduce two adaptive variants, TTT3R~\citep{chen2026ttt3r} and FILT3R~\citep{jin2026filt3r}, that we mainly focus on in this paper.
Appendix~\ref{app:additional_adaptive} formulates the additional adaptive
references, MeMix, TTSA3R, and ReCal3R.
\\
\textbf{TTT3R}~\citep{chen2026ttt3r} derives a separate coefficient for each CUT3R
scene token from its attention to the current observation. In the released
implementation, for $t\ge2$,
\begin{equation}
    a_{t,i}
    =\sigma\!\left(\operatorname{mean}\!\left(B_{t,i,:}\right)\right),
    \label{eq:ttt3r_coefficient}
\end{equation}
where $\sigma$ is the sigmoid and $B$ contains scaled, pre-softmax state-to-image attention scores.
The mean reduction averages across dimensions including decoder layers, attention heads, and
observation tokens, retaining the scene-token index only for each time step.
The first frame receives a full update, $a_{1,i}=1$.
Since the state-to-observation attention requires state tokens,
TTT3R is deeply coupled with CUT3R's state modeling, thus not directly applicable to ZipMap.
\\
\textbf{FILT3R}~\citep{jin2026filt3r} determines token-wise update weights for CUT3R
using a Kalman-style filter. It measures drift between successive
\emph{candidate} states,
$d_{t,i}=\|\widetilde S_{t,i}-\widetilde S_{t-1,i}\|_2$,
and maps its magnitude, normalized by a shared exponential moving
average, to process noise $Q_{t,i}$. Under the evaluated default
settings, the update weight for $t\ge2$ is the Kalman gain:
\begin{equation}
    a_{t,i}=\frac{P^-_{t,i}}{P^-_{t,i}+r},
    \label{eq:kalman}
\end{equation}
where $P^-_{t,i}$ is the predicted variance and $r$ is the
fixed measurement-noise parameter. Larger normalized drift increases
the gain for a given prior variance. The first frame receives a full
update, $a_{1,i}=1$. Please refer to FILT3R for the drift-to-noise mapping, variance
recursion, initialization, and implementation details.
Unlike TTT3R, 
FILT3R is a high-level filtering idea that can be applied to different backbones. 
Appendix~\ref{app:filt3r_zipmap} gives the update rule and filter settings for
our ZipMap adaptation, FILT3R(Z).
\\
\textbf{Discussion}
\label{ada_discussion}
Relative to vanilla updates ($a_t=1$), the adaptive variants
above both reduce the average update weight and vary it across
frames or state components. Their improvements therefore do not
isolate the benefit of adaptive modulation from that of conservative
updating. We ask: \emph{How much improvement remains when modulation
is removed while the average update weight is preserved?}

\section{How Much Does Adaptation Help?}
\label{sec:bootcon}

The question raised above
suggests a direct experiment: replace adaptive weights with a constant
while preserving their post-initialization mean.
The less the improvement remains, 
the more adaptive modulation contributes to the improvement,
and vice versa.
Next, we construct this control and examine its performance.

\subsection{A Constant Update Weight Control}
\label{sec:bootcon_rule}
We introduce \textbf{Bootstrap Once, Then Update Conservatively
(BootCon)} as the control. Starting from the pretrained initial state $S_0$, BootCon commits 
the first update in full and uses a small, constant weight thereafter:
\begin{equation}
    S_t=S_{t-1}+c_t\Delta_t,
    \qquad
    c_t=
    \begin{cases}
        1, & t=1,\\
        c, & t\ge2,
    \end{cases}
    \qquad 0\le c\le1.
    \label{eq:bootcon}
\end{equation}
The first write initializes a scene-conditioned state. We retain the same 
full-first-update convention used by TTT3R and FILT3R, rather
than introducing a different initialization only for BootCon. Subsequent
writes use the same scalar $c$ across all controlled memory components.

Under Equation~\ref{eq:adaptive}, BootCon simply sets
$a_t=c_t$. The rule therefore applies to both CUT3R's scene
tokens and streaming ZipMap's fast weights, preserving their proposal
computations and inference speed. Although the update \emph{weight} is
constant, the proposal $\Delta_t$ remains dependent on the current observation
and state. Setting $c=1$ recovers the vanilla update rule.

\subsection{Matching the Average Update Weight}
\label{sec:bootcon_calibration}
For fair comparison, we calibrate the constant weight $c$ to match 
the average update weight of the adaptive references.
Let $\mathcal C$ be the sequences in one benchmark and 
$T_s$ the number of frames
in sequence $s$. We define
\begin{equation}
    c_{\mathrm{match}}
    =\frac{1}{|\mathcal C|}
      \sum_{s\in\mathcal C}\frac{1}{T_s-1}
      \sum_{t=2}^{T_s}
      \operatorname{mean}\!\left(a^{\mathrm{ref}}_{s,t}\right),
    \label{eq:bootcon_calibration}
\end{equation}
where the inner mean averages over the memory components.
We exclude the full first update. Each adaptive reference produces its own
constant $c_{\mathrm{match}}$.
Calibrations neither use ground-truth information nor optimize
reconstruction scores. They are specified for fair comparison, 
as an in-benchmark diagnostic control rather than held-out validation.

\subsection{What Remains after Removing Adaptation?}
\label{sec:bootcon_matched_results}

We compare the vanilla backbone, an adaptive reference, and its
mean-matched BootCon control. The evaluation covers
camera-pose estimation, video-depth prediction, and 3D reconstruction.
Within each comparison, we keep the checkpoint, input sequence,
preprocessing, numerical precision, and scoring protocol unchanged.
Each sequence runs continuously, without intermediate memory resets or
overlapping-window alignment during inference.
For an error metric $E$, when the adaptive reference improves over vanilla
($E_{\mathrm{adaptive}} < E_{\mathrm{vanilla}}$), we report the
percentage of improvement recovered by BootCon as
\begin{equation}
    \mathrm{RI}
    =100\,\frac{E_{\mathrm{vanilla}}-E_{\mathrm{BootCon}}}
                   {E_{\mathrm{vanilla}}-E_{\mathrm{adaptive}}}.
    \label{eq:bootcon_recovered_improvement}
\end{equation}
An RI of $100\%$ indicates equal performance, and values above $100\%$
favor BootCon. When the adaptive reference does not improve over vanilla,
RI is inapplicable and we display a dash.
We report the original scores alongside this ratio in
Table~\ref{tab:bootcon_mean_matched} to make the remaining performance
differences explicit. Appendix~\ref{app:bootcon_full} provides the evaluation
settings and additional camera-pose metrics.
\input{tables/mean_matched_comparison}
\\
\textbf{Observation.}
Table~\ref{tab:bootcon_mean_matched} shows that removing adaptive
modulation preserves most of the improvement over vanilla backbones.
Across the evaluated tasks' primary metrics, mean-matched BootCon recovers
$93.9\%$--$123.6\%$ of the improvement achieved by the adaptive references.
Thus, much of the observed benefit can be reproduced by
conservative constant updating, without varying weight across
observations or memory components. This leads to the next question
\emph{why conservative updating is so effective}.
\section{Stability and Plasticity Probing}
\label{sec:diagnostics}

The mean-matched results suggest that reducing update weight explains
much of the adaptive references' accuracy gain. We investigate this
phenomenon by examining how constant and adaptive updates affect the
memory over time. Each update can both change the coordinate frame
supporting earlier predictions and incorporate information useful for
reconstructing incoming observations. We therefore use frozen-memory
probes to measure \emph{historical stability}, the preservation of earlier
predictions, and \emph{update-induced plasticity}, the improvement in a
window's reconstruction after its updates.
\input{figures/stability_plasticity}
\input{figures/stability_reprediction}
\\
\textbf{Stability}: \emph{does the later memory still reproduce its earlier predictions?}
We retain the original camera positions emitted for the first 64 frames.
After these frames and after every additional 64 frames, we use the
frozen current memory to reconstruct those same first 64 frames again.
We compare the re-predicted camera positions directly
with the saved positions, \emph{without fitting an alignment}.
We convert each probe's normalized
discrepancy into a consistency score between zero and one.
One means exact reproduction.
Lower stability scores mean greater historical drifting.
Figure~\ref{fig:stability_reprediction} (upper row) shows how this drift appears in one
scene. By frame 800, the re-predicted camera path and doorway have shifted
away from their original positions under vanilla CUT3R.
See Appendix~\ref{app:diagnostic_definitions} for the formal stability definition.\\
\textbf{Plasticity}: \emph{does learning from a window improve its reconstruction?}
We always evaluate the same final 64 frames of each sequence.
For each context length, we start a fresh stream from the corresponding
earlier frame, process the preceding context, and then learn from the
target window. We reconstruct that window twice, using frozen memory
from immediately before and after its updates.
We align each predicted trajectory to its ground truth \emph{separately} 
and measure its error by ATE.
Plasticity is the percentage by which the window's error decreases after
the updates. A value of $+20\%$ means a $20\%$ error reduction; zero means
no change, and a negative value means the updates make reconstruction
worse. Negative values are retained, and small pre-update errors can
amplify the percentage.
See Appendix~\ref{app:diagnostic_definitions} for the formal plasticity definition.
\input{tables/context_ratio_comparison}
\\
\textbf{Observations.}
We compare vanilla CUT3R, three adaptive methods (TTT3R, FILT3R,
and TTSA3R), and their mean-matched BootCon controls using the above
probes on ScanNet, with detailed settings in Appendix~\ref{app:diagnostic_setup}.
\\
1) \emph{Stability is closely reproduced.}
In Figure~\ref{fig:diagnostics}, each calibrated BootCon curve follows
its adaptive reference's declining stability trend. The controls retain
$101.5\%$, $95.0\%$, and $99.0\%$ of the stability of TTT3R, FILT3R,
and TTSA3R, respectively.
Thus, retaining the adaptive mean reproduces much of the observed
historical stability.
\\
2) \emph{Conservative updates preserve useful plasticity.}
Vanilla CUT3R exhibits negative mean plasticity at eight of the fifteen
tested context lengths, indicating that memory writes can worsen
reconstruction of newly observed frames. TTT3R and TTSA3R reduce the
frequency and severity of these negative responses, but do not eliminate
them. FILT3R and its mean-matched BootCon control maintain positive mean
plasticity across all tested contexts, showing that conservative constant
updates can preserve learning capability without adaptive modulation.
\\
\textbf{Interpretation.}
Our intuition is that each memory write incorporates useful scene
information together with estimation noise. Stronger writes give both a
larger influence on the state. As writes accumulate, their errors can
shift the spatial reference supporting earlier predictions. Meanwhile,
noisy or redundant information can interfere with useful structure in the
fixed-size memory, bringing it to its effective capacity limit sooner and
making later writes less beneficial. Conservative updating can ease both
pressures by limiting the update weight assigned to each observation. Smaller writes can reduce historical drift and the buildup of
interference while retaining useful contributions from incoming
observations. The joint stability and plasticity results are consistent
with this account, although the probes do not directly measure noise
accumulation or memory saturation.
\\
\textbf{Beyond Mean Matching.}
Mean matching reproduces much of the adaptive methods' behavior, but does
not establish whether the matched update weight is well chosen. We
therefore move beyond reproducing adaptive methods and ask:
\emph{how to choose an effective update weight $c$ directly?}
\input{figures/camera_trajectories}

\section{\texorpdfstring{BootCon ($c_{\mathrm{ctx}}$):\\Choosing $c$ from the Context Ratio}{BootCon (cctx): Choosing c from the Context Ratio}}
\label{sec:bootcon_selection}
We take the checkpoint's nominal training context as a reference budget
for accumulated update weight. Choosing $c$ then amounts to deciding
how to spread this budget over the intended inference horizon.
The principle is simple: \textbf{longer streams, smaller updates.}
We hypothesize that the optimal update weight scales as
\begin{equation}
    c^\star \propto c_{\mathrm{ctx}}, \qquad
    c_{\mathrm{ctx}} = \frac{L_{\mathrm{train}}}{L_{\mathrm{infer}}},
    \label{eq:bootcon_context_selection}
\end{equation}
where $c^\star$ denotes the optimal post-bootstrap update weight for the
given backbone and task. Here $L_{\mathrm{train}}$ is the
checkpoint's nominal training context length, and
$L_{\mathrm{infer}}$ is the intended inference horizon, both measured in
processed frames. We use $c=c_{\mathrm{ctx}}$ as a simple default and retain
a full first update. At an 800-frame horizon, CUT3R's 64-frame training
context gives $c_{\mathrm{ctx}}=0.08$, while streaming ZipMap's 24-frame
context gives $c_{\mathrm{ctx}}=0.03$.

\subsection{Performance}
\textbf{Stability and plasticity.}
Using the same probe setting in Section~\ref{sec:diagnostics}, 
Figure~\ref{fig:diagnostics} shows that BootCon($c_{\mathrm{ctx}}$) achieves the highest
mean stability at every tested context length. The second row of
Figure~\ref{fig:stability_reprediction} provides a qualitative illustration:
as context grows, the re-predicted camera path and doorway remain closer
to their original positions than under vanilla CUT3R.
In Figure~\ref{fig:diagnostics}, the context ratio default achieves the
highest mean plasticity at most tested context lengths, outperforming
both the adaptive references and their mean-matched BootCon controls.
\\
\phantomsection
\noindent\textbf{Comparison with other methods.}
\label{sec:bootcon_context_comparison}
Table~\ref{tab:bootcon_context_comparison}
compares CUT3R, its seven adaptive variants (TTT3R, FILT3R, ReCal3R, MeMix,
TTSA3R, AFG-Img and AFG-Pose), streaming ZipMap, and BootCon($c_{\mathrm{ctx}}$) applied to each
backbone on TUM and ScanNet pose, Bonn and KITTI depth, and 7-Scenes and
NRGBD reconstruction. We first use a design horizon for each benchmark
(Table~\ref{tab:context_comparison_settings}, Appendix~\ref{app:bootcon_context_comparison}).
In this setting, ZipMap-based BootCon achieves the lowest primary error
on all six benchmarks. It reduces Bonn scale-aligned Abs Rel by $43\%$
relative to vanilla ZipMap. On 7-Scenes, both variants substantially
improve reconstruction accuracy and completeness over their backbones,
with ZipMap-based BootCon($c_{\mathrm{ctx}}$) achieving the lowest errors among the compared
methods.
However, BootCon($c_{\mathrm{ctx}}$) does not always achieve the best performance on every
metric. CUT3R-based BootCon worsens TUM and ScanNet RPE-R despite improving ATE.
Figure~\ref{fig:scannet_pose_qualitative} shows TUM and ScanNet camera trajectories,
with additional examples in Appendix~\ref{app:pose_qualitative}.
Figures~\ref{fig:reconstruction_teaser} and~\ref{fig:reconstruction_qualitative}
show qualitative reconstruction examples on 7-Scenes.
These results show that BootCon($c_{\mathrm{ctx}}$) sets a strong baseline
across tasks.
\input{figures/reconstruction_qualitative}
\\
\textbf{Performance across prefixes.}
Supplementary Figure~\ref{fig:prefix_benchmarks} tracks the eleven methods
from shorter prefixes to the endpoints in Table~\ref{tab:bootcon_context_comparison}.
Each BootCon curve uses one post-bootstrap weight throughout, set by the
benchmark-specific design horizon. Both variants improve on their respective
backbones at every measured prefix. Across the eleven methods, a BootCon variant has
the lowest error at all six endpoints and in 16 of the 17 intermediate
comparisons. The exception is 7-Scenes at 200 frames, where ReCal3R narrowly leads.
\\
\textbf{One design horizon across benchmarks.}
When stream lengths are unknown, we use the same 1,000-frame design horizon
on all six benchmarks. This fixes $c=L_{\mathrm{train}}/1000$ at $0.064$
for CUT3R and $0.024$ for streaming ZipMap. Each backbone uses its preset weight from the second
frame onward, with no benchmark-specific length statistics or tuning.
The shared-horizon rows in Table~\ref{tab:bootcon_context_comparison}
use the same scoring caps as the benchmark-specific rows. With this
single horizon, CUT3R-based BootCon has lower ScanNet ATE than all nine
comparison methods, and ZipMap-based BootCon outperforms them on the
primary metric of the other five benchmarks.
\\
\textbf{Coefficient sensitivity.}
To assess the context-ratio heuristic, we search around $c_{\mathrm{ctx}}$
by repeatedly halving or doubling the coefficient for each backbone and
dataset.
Across the six datasets in supplementary Figure~\ref{fig:bootcon_selection},
$c_{\mathrm{ctx}}$ matches the lowest tested error for CUT3R on TUM,
both backbones on ScanNet, and ZipMap on 7-Scenes. Smaller coefficients perform better on
Bonn and KITTI depth, but the context-ratio errors remain within $2.8\%$
of the respective tested minima. Overall, this holds for 10 of the 12
backbone--dataset combinations.
Thus, the ratio provides a useful starting point, not a universal
optimum; the best coefficient depends on the backbone, dataset, and task.
These sweeps are in-benchmark sensitivity analyses, not held-out
validation or a separate refinement stage.
\input{tables/bootstrap_ablation}
\\
\phantomsection
\noindent\textbf{Ablations.}
\label{sec:bootcon_bootstrap_ablation}
Table~\ref{tab:bootcon_bootstrap_ablation} ablates ZipMap-based BootCon's bootstrap and
subsequent update schedule on TUM. 
In (a), we vary initialization while
fixing later updates at $c_{\mathrm{ctx}}=0.03$. Omitting the full bootstrap
raises ATE by $48\%$, proving the bootstrap is necessary. However, too strong bootstrap (tripling its weight) also worsens ATE.
 In (b), we retain a
full bootstrap and match the post-bootstrap mean at $0.03$. Periodic and
randomly placed full writes, with one fractional write for exact budget
matching, yield about $3\%$ higher mean ATE than default BootCon. Random
framewise coefficients, drawn uniformly and rescaled to the same
post-bootstrap mean, yield nearly the same ATE as default BootCon.
These results show that BootCon is robust to a series of variations.
\section{Conclusion}
\label{sec:conclusion}

BootCon simplifies online 3D reconstruction with state by committing the
first update in full and applying a small, constant weight thereafter.
Across CUT3R and streaming ZipMap on six benchmarks, mean-matched BootCon
recovers at least $94\%$ of the adaptive references' improvement on the
reported primary metrics, sometimes outperforming them.
Stability--plasticity probes on ScanNet further show that the mean-matched
controls broadly follow their adaptive references' patterns,
suggesting similar memory dynamics can be achieved without adaptive modulation. 
Beyond this control, the
training-to-inference context ratio selects an effective weight directly,
without running an adaptive reference. With benchmark-specific design
horizons, ZipMap-based BootCon outperforms all nine comparison methods
on the primary metric of each benchmark. With one 1,000-frame horizon,
BootCon retains this advantage on all six benchmarks.
These results establish BootCon as an effective practical approach.

We do not claim that BootCon is the optimal approach to online 3D
reconstruction. The context ratio is a useful default, and gains in global
accuracy can still coexist with worse local motion. We advocate developing
more effective adaptive methods and proposing more challenging benchmarks that demonstrate clear benefits beyond
this simple baseline, making the contribution of adaptive modulation
explicit.

\FloatBarrier
\section*{LLM Usage Statement}
OpenAI Codex assisted with drafting and revising text throughout the
manuscript and supplementary material, preparing LaTeX, tables, and
visualizations, developing and debugging experimental and evaluation code,
running experiments, and analyzing results. It also assisted with
literature discussions and developing and checking methodological
explanations. The authors directed the study and retain responsibility for the final submission,
including its text, code, analyses, and scientific claims.

\begingroup
\small
\bibliographystyle{ieeenat_fullname}
\bibliography{references}
\endgroup

\clearpage
\maketitlesupplementary
\raggedbottom
\appendix
\section{Mean-Matched Evaluation and Additional Pose Results}
\label{app:bootcon_full}

\subsection{Evaluation settings}
\label{app:bootcon_protocol}
For Tables~\ref{tab:bootcon_mean_matched} and~\ref{tab:bootcon_rpe_matched},
we use the same backbone checkpoints, input resolutions, and scoring
procedures as in Appendix~\ref{app:bootcon_context_comparison}.

We calibrate on RGB alone, using up to 1,000 frames per sequence for
pose and 7-Scenes and 500 for video depth and NRGBD. Following
Equation~\ref{eq:bootcon_calibration}, we average post-bootstrap update
weights within each sequence, average equally across sequences, and
round to two decimals. For each backbone and adaptive reference, we use
this constant at every prefix of the benchmark. On ScanNet, all 100 RGB
streams enter calibration. At each scoring prefix, all methods use the
same subset with finite ground-truth poses throughout the evaluated prefix.

\subsection{Additional camera-pose metrics}
\label{app:bootcon_tables}
Replacing adaptive update weights with their mean generally preserves
the accuracy of estimated camera motion between consecutive frames
(Table~\ref{tab:bootcon_rpe_matched}). On TUM at 800 frames, both
mean-matched controls lower translation and rotation errors. On ScanNet
at 1,000 frames, the TTT3R control matches the reported translation error
and lowers rotation error, while the FILT3R control increases both.

\input{tables/mean_matched_relative_pose}

\subsection{Additional camera trajectories}
\label{app:pose_qualitative}
Figure~\ref{fig:additional_pose_qualitative} adds two trajectories to
Figure~\ref{fig:scannet_pose_qualitative}. We align each trajectory to
ground truth over the full prefix with Sim(3), measure ATE in 3D, and plot
all methods on the principal plane of the ground-truth trajectory for each scene.

\input{figures/camera_trajectories_additional}
\section{Historical Stability and Relative Plasticity}
\label{app:diagnostics}

This appendix gives the exact definitions used in
Figure~\ref{fig:diagnostics} and Section~\ref{sec:diagnostics}.
Both axes score frozen-memory re-queries at individual context lengths.

\subsection{Sequences, Prefixes, and Probe Times}
\label{app:diagnostic_setup}

We use the 14 prepared ScanNet sequences with at least 1,000 sampled
frames and finite GT poses throughout the first 1,000, at sampling
stride 3. Their identifiers are \texttt{scene0712\_00},
\texttt{scene0721\_00}, \texttt{scene0739\_00}, \texttt{scene0744\_00},
\texttt{scene0747\_00}, \texttt{scene0752\_00}, \texttt{scene0753\_00},
\texttt{scene0754\_00}, \texttt{scene0756\_00}, \texttt{scene0757\_00},
\texttt{scene0766\_00}, \texttt{scene0776\_00}, \texttt{scene0785\_00},
and \texttt{scene0793\_00}. All methods use this same fixed cohort,
with $T=1000$ and target-window size $w=64$. Frame counts below refer to
sampled frames, not raw video frames. This diagnostic subset is distinct
from the larger pose-benchmark cohort.

We calibrate the matched controls on these streams without rounding and
average both stability and plasticity scores equally across sequences.
Context lengths range from 0 to 896 in steps of 64: subsequent frames for
stability, preceding frames for plasticity. Using one-based inclusive
indexing, the context grid and the two fixed target windows are
\begin{equation}
 \begin{aligned}
 \mathcal C&=\{0,64,\ldots,896\},\\
 H&=\{1,\ldots,w\},\qquad B=\{T-w+1,\ldots,T\}.
 \end{aligned}
 \label{eq:diagnostic_prefix_windows}
\end{equation}
For stability, run one uninterrupted stream from frame 1 and query $H$
after frame $w+C$ for each $C\in\mathcal C$. The last probe is at frame
960; the remaining 40 frames do not enter the plotted score.
For plasticity, start a fresh state and adaptive history for every $C$,
stream frames $T-w-C+1,\ldots,T-w$, then process $B$ (frames 937--1000).
At $C=0$ there are no preceding frames; at $C=896$ the stream starts at
frame 41. No frames are padded, mirrored, or repeated. Thus $C$ counts
\emph{subsequent} context for stability and \emph{preceding} context for
plasticity, excluding the 64 target frames in either panel. The targets
remain fixed within each scene, but increasing context introduces more
content; this is not a strictly content-controlled write-count experiment.

\subsection{Exact Definitions}
\label{app:diagnostic_definitions}

\paragraph{Historical stability against emitted predictions.}
For each method, let $h_{s,i}\in\mathbb R^3$ be the camera position
originally emitted for frame $i\in H$ of sequence $s$. Let $q_{s,C,i}$ be the position
re-predicted for that same frame from the frozen state after frame $w+C$.
The first-window radius and normalized positional drift are
\begin{align}
 \bar h_s &= \frac1w\sum_{i=1}^w h_{s,i},\notag\\
 r_s &= \sqrt{\frac1w\sum_{i=1}^w\|h_{s,i}-\bar h_s\|_2^2},\\
 d_{s,C} &=
 \frac{\sqrt{\frac1w\sum_{i=1}^w\|q_{s,C,i}-h_{s,i}\|_2^2}}
 {\max(r_s,10^{-12})}.
\end{align}
Convert each scene's current drift to a consistency score, then average
these scores equally across scenes:
\begin{equation}
 S_s(C)=\frac{1}{1+d_{s,C}},
 \qquad
 S(C)=\frac1{|\mathcal S|}\sum_{s\in\mathcal S}S_s(C),
 \quad |\mathcal S|=14.
 \label{eq:prefix_historical_stability}
\end{equation}
There is no rigid or similarity alignment, and neither trajectory is
recentered for the positional comparison. Centering is used only to
compute the reference radius. Ground truth is absent from the score.
A value of one means exact positional reproduction. The score does not
include camera-rotation error, and it combines changes of coordinate
frame with changes in reconstructed geometry. At $C=0$, we re-predict the
first 64 camera positions using the memory after frame 64. Updates within
this window may already have shifted earlier predictions, so $S(0)$ can
be below one before any additional context arrives.

\paragraph{Relative plasticity from pre/post window errors.}
For sequence $s$ and context length $C$, we query the same 64 target
frames using the memory before and after the window's updates.
Let $e^{\mathrm{pre}}_{s,C}$ and $e^{\mathrm{post}}_{s,C}$ be the
corresponding local ATEs, each computed against ground truth after
independent Sim(3) alignment.

The plotted plasticity is the equal-scene mean percentage reduction for
this fixed target and context length:
\begin{align}
 p_s(C)&=100\,
 \frac{e^{\mathrm{pre}}_{s,C}-e^{\mathrm{post}}_{s,C}}
      {e^{\mathrm{pre}}_{s,C}+\epsilon},
 \qquad \epsilon=10^{-8}\ \mathrm{m},\\
 \mathcal P(C)&=\frac1{|\mathcal S|}\sum_{s\in\mathcal S}p_s(C).
 \label{eq:prefix_relative_plasticity}
\end{align}
Positive values indicate reduced local ATE, zero indicates no change,
and negative values indicate increased error. We do not clip negative
values or discard windows with small errors. Ratios can be sensitive
to a small pre-update denominator.

\subsection{Read-Only Queries and Compared Settings}

Each probe image independently reads a frozen snapshot, leaving scene
memory, pose memory, and adaptive histories unchanged.

For the mean-matched BootCon controls in Figure~\ref{fig:diagnostics},
we average each adaptive reference's
weights over frames 2--1,000 within each sequence, then equally across
sequences, using RGB alone. This gives $c=0.27410882377032636$ for TTT3R,
$0.1522137875484942$ for FILT3R, and $0.2628284443435773$ for TTSA3R.
\FloatBarrier
\section{Context-Based Coefficient Selection: Protocols and Additional Results}
\label{app:bootcon_selection}

\subsection{Coefficient Sensitivity}
\label{app:selection_protocols}
For each backbone and benchmark, we evaluate five weights obtained by
multiplying $c_{\mathrm{ctx}}$ by $\{1/4,1/2,1,2,4\}$. On TUM and Bonn, we
test additional weights to better characterize the low-error region.

We compare candidates with the same settings as
Table~\ref{tab:bootcon_context_comparison}, using the scoring protocols in
Appendix~\ref{app:comparison_metrics}.

\input{figures/context_ratio_sweep}

Figure~\ref{fig:bootcon_selection} shows ATE for pose, scale-aligned Abs Rel
for depth, and Chamfer distance for reconstruction. Stars mark
$c_{\mathrm{ctx}}$.
Within the five-point Bonn grid, CUT3R and ZipMap attain their lowest Abs Rel at
$c_{\mathrm{ctx}}/4$ and $c_{\mathrm{ctx}}/2$, respectively.
On 7-Scenes, both backbones attain their lowest reconstruction accuracy
error (Acc) at $c_{\mathrm{ctx}}/2$ within the five-point grid.

\subsection{Datasets and Prefix Caps}
\label{app:comparison_datasets}

All methods process the same frames in order, without resets, replay,
loop closure, or online window alignment. Each evaluated prefix length
is a cap. Shorter streams contribute through their final frame without
padding.

\paragraph{Pose Benchmarks: TUM and ScanNet.}
We follow the pose evaluation of TTT3R and FILT3R~\citep{chen2026ttt3r,jin2026filt3r}.
TUM uses all eight Freiburg3 sitting and walking sequences at stride 1,
up to 800 frames. ScanNet uses stride 3 and the same 67 streams with
finite ground-truth poses throughout the capped stream for every method
and prefix. Only 14 reach the cap of 1000 processed frames.

\paragraph{Depth Benchmarks: Bonn and KITTI.}
Both benchmarks use stride 1 and a cap of 500 frames.
Following TTT3R~\citep{chen2026ttt3r} and FILT3R~\citep{jin2026filt3r},
we start evaluation at frame 31 of each Bonn sequence.
The five evaluated sequences are
\texttt{balloon2}, \texttt{crowd2}, \texttt{crowd3},
\texttt{person\_tracking2}, and \texttt{synchronous}. KITTI uses all
13 validation clips.

\paragraph{Reconstruction Benchmarks: 7-Scenes and NRGBD.}
We use all 18 test sequences from 7-Scenes, including stairs, and all
nine prepared NRGBD sequences, at stride 1. The caps are 1000 and 500
frames, respectively.

\subsection{Evaluation Metrics}
\label{app:comparison_metrics}

\paragraph{Pose Metrics.}
We compute ATE as
positional RMSE after one positive-scale, reflection-free Sim(3)
alignment over each evaluated trajectory. RPE-T and RPE-R measure
translation and rotation errors between consecutive frames.
Table~\ref{tab:bootcon_context_comparison} reports ATE and RPE-T in meters
and RPE-R in degrees, averaging each metric equally across sequences.

\paragraph{Depth Metrics.}
Following FILT3R~\citep{jin2026filt3r}, we align depth by one scale
factor per sequence prefix. Abs Rel averages the absolute depth error
at each valid pixel divided by its ground-truth depth. Threshold accuracy
($\delta<1.25$) is the percentage of valid pixels whose predicted depth
lies strictly between $0.8$ and $1.25$ times the ground-truth depth.
Log RMSE is the root mean squared difference between predicted and
ground-truth log depths. Lower Abs Rel and log RMSE and higher threshold
accuracy are better. We weight each sequence's score by its valid-pixel count.

\paragraph{Reconstruction Metrics.}
We follow CUT3R's reconstruction protocol~\citep{wang2025cut3r} using the
scoring routines released with FILT3R~\citep{jin2026filt3r}. We normalize
scale and depth shift over each prefix, apply a central $224\times224$ crop,
and align the valid points by rigid ICP with a $0.1$\,m threshold.
We compute exact nearest-neighbor distances over all valid points, summarize
the pointwise scores by their mean and median within each scene, then
average equally across scenes. Acc and Comp are
in meters; NC is dimensionless. Chamfer distance averages mean Acc and
mean Comp.

\subsection{Evaluation Settings for Table~\ref{tab:bootcon_context_comparison}}
\label{app:bootcon_context_comparison}

For Table~\ref{tab:bootcon_context_comparison}, we use the released CUT3R
4--64-view and streaming ZipMap checkpoints with the datasets and scoring
protocols in Appendices~\ref{app:comparison_datasets} and~\ref{app:comparison_metrics}.

\input{tables/context_ratio_settings}
\input{figures/prefix_benchmarks}

\paragraph{Context lengths and update weights.}
Table~\ref{tab:context_comparison_settings} gives the settings used in
Section~\ref{sec:bootcon_selection}. We use training contexts of 64
processed frames for CUT3R and 24 for streaming ZipMap. The first write
has weight one, and subsequent writes use
$c_{\mathrm{ctx}}=L_{\mathrm{train}}/L_{\mathrm{infer}}$.
For the benchmark-specific rows in Table~\ref{tab:bootcon_context_comparison}
and the curves in Figure~\ref{fig:prefix_benchmarks}, the design horizon
$L_{\mathrm{infer}}$ equals the median capped sequence length in processed
frames, rounded up to a multiple of 100.
It equals the scoring cap except on ScanNet pose (600 versus 1000) and
KITTI depth (200 versus 500). The shared-horizon rows use
$L_{\mathrm{infer}}=1000$ with the same sequences and scoring caps.

\section{Backbone and Adaptive-Update Details}
\label{app:state_updates}

\subsection{ZipMap Streaming Updates}
\label{app:zipmap_streaming}

We use ZipMap's released streaming checkpoint~\citep{jin2026zipmap},
with one write followed by one application per incoming frame.
All pretrained feature projections and prediction heads remain frozen.

\paragraph{Write proposal.}
Each TTT layer and head has three fast-weight matrices, collected as
$\mathcal W$, defining a gated MLP
\begin{equation}
 f_{\mathcal W}(k)
 =\bigl[\operatorname{SiLU}(kW^{(0)})\odot(kW^{(2)})\bigr]W^{(1)}.
\end{equation}
The frozen network produces query, key, and value tokens and positive,
token-wise learned rates $\ell^{(a)}_{t,i}$ for matrix $a$.
For the current frame's tokens $\mathcal I_t$, the native Hebbian write is
\begin{equation}
 \begin{aligned}
 H_t^{(a)}&=
 \sum_{i\in\mathcal I_t}\ell^{(a)}_{t,i}
 \nabla_{W^{(a)}}
 \bigl\langle f_{\mathcal W_{t-1}}(k_{t,i}),v_{t,i}\bigr\rangle,\\
 G_t^{(a)}&=\operatorname{NS}\!\left(H_t^{(a)}\right),
 \end{aligned}
 \label{eq:zipmap_native_write_details}
\end{equation}
where $\operatorname{NS}$ is the native Newton--Schulz matrix transform
used by Muon. Thus $G_t$ in Section~\ref{sec:prelim} denotes the
\emph{post-Muon direction}, not the raw Hebbian gradient $H_t$.
The positive sign implements ascent on the key--value inner product.
The learned rates remain unchanged in all our controls. BootCon and
FILT3R(Z) instead scale the resulting direction \emph{after} Muon.

\paragraph{Commit and apply.}
Suppressing the matrix, layer, and head indices, the commit is
\begin{equation}
 W_t=\mathcal N_t(W_{t-1}+a_tG_t),
 \qquad
 \mathcal N_t(X)_{:,j}
 =n_j\frac{X_{:,j}}{\|X_{:,j}\|_2+10^{-5}},
 \label{eq:zipmap_native_normalization_details}
\end{equation}
where $n_j$ is the native cached reference norm. These are column
vectors in the unfused kernel's matrix layout; an equivalent transposed
layout normalizes rows. The rule preserves their reference magnitudes,
up to the numerical stabilizer, rather than imposing unit norms or an
angle cap. Vanilla ZipMap uses $a_t=1$; BootCon uses $a_1=1$ and
$a_t=c$ thereafter. The current query tokens are then evaluated with
the committed $\mathcal W_t$, followed by the native output normalization
and projection. Consequently, unlike CUT3R's scene-state commit, this
weight affects both the current prediction and the memory carried
to future frames.

\subsection{Additional Adaptive Variants}
\label{app:additional_adaptive}

The following formulations describe the released CUT3R-based variants
used in our additional comparisons. All modify the scene-token
coefficient in Equation~\ref{eq:adaptive}; the auxiliary pose-retrieval
memory retains its native update. We preserve each reference's
initialization behavior rather than imposing a shared first-frame rule.

\paragraph{AFG.}
Both AFG variants~\citep{ren2026afg} scale the TTT3R token gate using
the released settings on every dataset. AFG-Img uses changes in pooled
image features (threshold 1.5), and AFG-Pose uses changes in pose features
(threshold 1.0, sigmoid sharpness 3.0).

\paragraph{MeMix: binary token selection.}
The evaluated plain MeMix variant~\citep{dong2026memix} scores each
candidate scene token against the mean projected image feature
$\bar x_t$ and commits only the $k$ lowest-scoring tokens:
\begin{equation}
 s_{t,i}=\langle\widetilde S_{t,i},\bar x_t\rangle,
 \qquad
 a_{t,i}=\mathbf 1\!\left[i\in\operatorname{BottomK}_k(s_t)\right].
 \label{eq:memix_commit_details}
\end{equation}
Its released setting uses $k=708$ of $N=768$ one-token groups, without
additional soft gates. Hence the frame-mean weight is exactly
$708/768=0.921875$, although the selected tokens change with the input.
This binary selection also applies to the first frame: native MeMix
does \emph{not} perform the full first update used by BootCon.

\paragraph{TTSA3R: temporal and spatial gates.}
For $t\ge2$, TTSA3R~\citep{zheng2026ttsa3r} combines candidate-state
drift with image-feature change. Let
$d_{t,i}=\|\widetilde S_{t,i}-\widetilde S_{t-1,i}\|_2$,
$\bar d_t=N^{-1}\sum_i d_{t,i}$, and let $x_{t,j}$ be encoder features
at image-token position $j$. The released coefficients are
\begin{align}
 u_{t,i}&=\sigma\!\left(d_{t,i}/\bar d_t-1.5\right),\\
 v_{t,i}&=\sigma\!\left(
   \max_j\left\{|\bar B_{t,ij}|\,
   [1-\cos(x_{t,j},x_{t-1,j})]\right\}\right),\notag\\
 a_{t,i}&=u_{t,i}v_{t,i}.
 \label{eq:ttsa3r_commit_details}
\end{align}
Here $\bar B$ averages the pre-softmax attention scores over decoder
layers and heads before taking the absolute value; the pose-token
column is excluded. The feature comparison uses corresponding image-token
positions, not an estimated optical flow. The first update is full,
$a_{1,i}=1$.

\paragraph{ReCal3R: reliability-calibrated conservative writing.}
ReCal3R~\citep{li2026recal3r} starts from an attention-derived coefficient
$\beta_{t,i}$. It attenuates this coefficient using a feature-residual
factor $r_t$ and an EMA of past update coefficients, $u_{t,i}$:
\begin{equation}
 u_{t,i}=0.95u_{t-1,i}+0.05\beta_{t,i},
 \qquad b_{t,i}=\beta_{t,i}\,r_t\exp(-u_{t,i}).
\end{equation}
The residual factor is the sigmoid of the relative difference between
projected input and decoded image features, offset by $0.2$.
A token reliability $\gamma_{t,i}$ combines rank-normalized drift from
the initial state with rank-normalized attention entropy. The released
fallback weight and final commit are
\begin{align}
 R_{t,i}&=1-\gamma_{t,i}
       [1-4\gamma_{t,i}(1-\gamma_{t,i})],\\
 b^{\mathrm{mix}}_{t,i}&=(1-R_{t,i})b_{t,i}+R_{t,i}b_0,\\
 a_{t,i}&=\operatorname{clip}_{[0,1]}\!\left(
   \tfrac12 b^{\mathrm{mix}}_{t,i}
   +\tfrac12\operatorname{mean}_j b^{\mathrm{mix}}_{t,j}\right),
 \qquad b_0=0.1.
 \label{eq:recal3r_commit_details}
\end{align}
The first update is full. We execute the released implementation,
including its rank statistics and smoothing, and calibrate BootCon
against the \emph{final applied} $a_{t,i}$ for $t\ge2$, not against the
internal base rate $b_0$.

\section{Adaptation of FILT3R to ZipMap}
\label{app:filt3r_zipmap}
Our \emph{FILT3R(Z)} applies FILT3R's filtering principle to ZipMap's
fast-weight writes. This is our adaptation, not an official FILT3R
implementation for ZipMap. It retains the native proposal, per-frame
update order, and normalization described in
Appendix~\ref{app:zipmap_streaming}.

\paragraph{Candidate and drift.}
For each fast-weight matrix, layer, and head, form an uncommitted
\emph{full-update} candidate using the post-Muon direction $G_t$:
\begin{equation}
    \begin{aligned}
    \widetilde W_t&=\mathcal N_t(W_{t-1}+G_t),\\
    d_t&=\operatorname{RMS}(\widetilde W_t-\widetilde W_{t-1}),\\
    \operatorname{RMS}(X)&=\sqrt{\frac{1}{|X|}\sum_{ij}X_{ij}^2}.
    \end{aligned}
    \label{eq:zipmap_filter}
\end{equation}
The comparison is between successive candidates, not between the
candidate and its own input state. Matrix RMS replaces CUT3R's
token-vector drift, producing one scalar weight per matrix and head.
Each such filter maintains its own previous candidate, drift EMA, and
variance; statistics are not pooled across layers or matrices.

\paragraph{Filter and commit.}
The first frame receives $K_1=1$, and we initialize $P_1=1.5$.
The first measurable drift initializes $m_2=\max(d_2,\delta)$.
For subsequent frames, update the EMA \emph{before} normalizing drift:
\begin{align}
 m_t&=\max\!\left((1-\rho)m_{t-1}+\rho d_t,\delta\right)
       \quad(t\ge3),\\
 z_t&=\frac{d_t}{m_t+\epsilon},\quad(t\ge2),\notag\\
 Q_t&=q_{\min}+(q_{\max}-q_{\min})\notag\\
 &\qquad{}\times\sigma\!\left(\kappa(z_t-\tau)\right)\quad(t\ge2),\\
 P^-_t&=P_{t-1}+Q_t,\notag\\
 K_t&=\operatorname{clip}_{[k_{\min},k_{\max}]}\!\left(
       \frac{P^-_t}{P^-_t+r+\epsilon}\right),\\
 P_t&=(1-K_t)^2P^-_t+rK_t^2,\notag\\
 W_t&=\mathcal N_t(W_{t-1}+K_tG_t).
 \label{eq:zipmap_filter_recursion}
\end{align}
The variance and commit recursions apply for $t\ge2$; the first commit
is the native full update. The saved candidate is always
$\widetilde W_t$, even when the committed weight is smaller than one.
As in BootCon, the weight scales $G_t$ \emph{before} normalization;
we do not interpolate the old state with an already normalized candidate.

\paragraph{Evaluated settings.}
The paper-default-filter comparisons use $q_{\min}=0.02$,
$q_{\max}=0.5$, $\kappa=20$, $\tau=3$, $r=1$, $\rho=0.05$,
$\delta=0.01$, $\epsilon=10^{-6}$, and
$(k_{\min},k_{\max})=(0.01,0.99)$. The first-frame override is outside
these clipping bounds. We measure the applied $K_t$ across matrices,
heads, and layers and then across all frames $t\ge2$ and sequences,
following Appendix~\ref{app:bootcon_protocol}, to obtain the matched
BootCon constant. Matching this mean removes the variation in weight
while retaining ZipMap's native, observation-dependent write proposal.

\end{document}

%% file: figures/teaser.tex
\begin{minipage}{\textwidth}
    \centering
    \includegraphics[width=\linewidth,height=0.30\textheight,keepaspectratio]{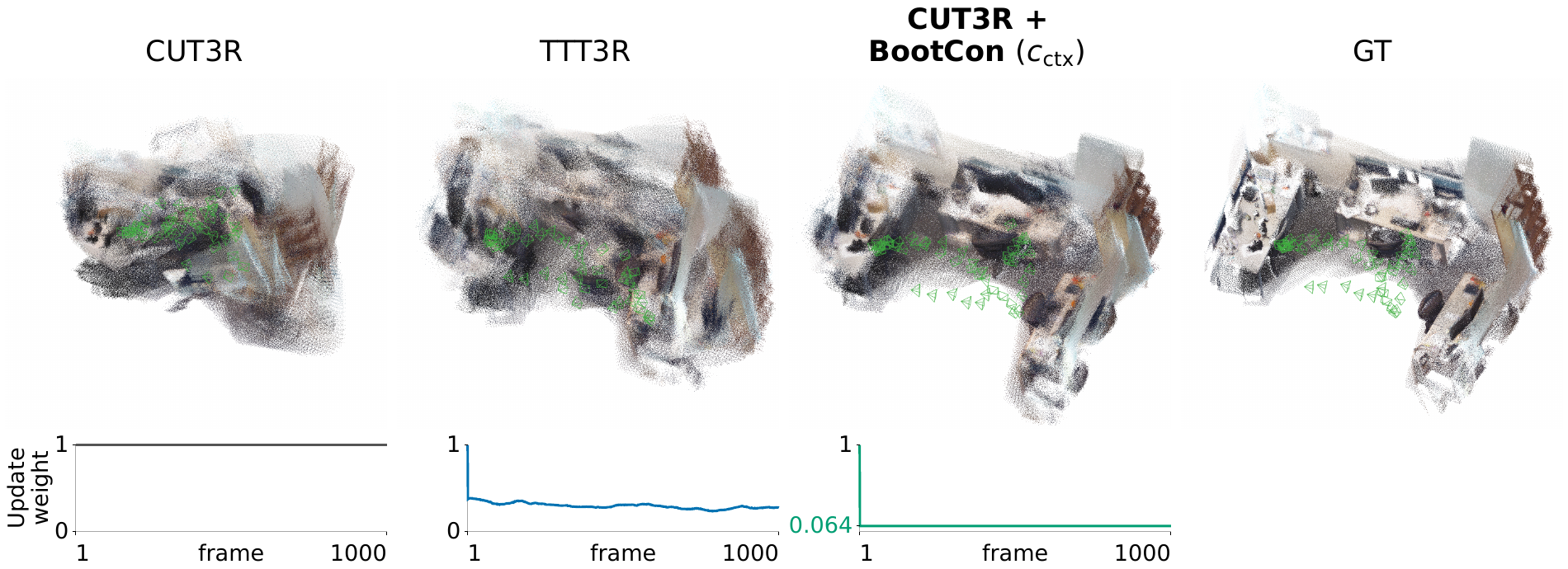}
    \captionof{figure}{\textbf{Bootstrap once, then update conservatively.}
    State-based online 3D reconstruction models maintain scene memory, updating
    it with each incoming image to predict camera poses and scene geometry.
    The update weight scales the model's predicted memory change: a weight
    of 1 applies the full update. BootCon applies one full update to initialize
    memory, then uses a small constant weight. Shown are reconstructions of
    the same 1,000-frame stream, with ground truth (GT) on the right;
    curves below each method show its update weight over time. With this simple fixed update rule, BootCon significantly outperforms more complex adaptive weight update method TTT3R.
    Additional examples appear in Figure~\ref{fig:reconstruction_qualitative}.}
    \label{fig:reconstruction_teaser}
\end{minipage}

%% file: tables/mean_matched_comparison.tex
\begin{table}[tp]
    \centering
    \small
    \setlength{\tabcolsep}{1.2pt}
    \renewcommand{\arraystretch}{1.0}
    \setlength{\aboverulesep}{0.75pt}
    \setlength{\belowrulesep}{0.75pt}
    \captionsetup{skip=3pt}
    \caption{\textbf{Adaptive References vs. Mean-Matched BootCon.}
    We compare vanilla backbones, adaptive methods, and their mean-matched BootCon. Mean-matched BootCon retains the full first update,
    then uses a constant $c_{\mathrm{match}}$ equal to each adaptive reference's
    mean update weight on that benchmark.
    RI reports the percentage
    of the adaptive method's improvement over vanilla recovered by BootCon:
    $100\%$ means full recovery, and values above $100\%$ favor BootCon.
    FILT3R(Z) is our adaptation of FILT3R to ZipMap
    (Appendix~\ref{app:filt3r_zipmap}). More metrics are in Appendix~\ref{app:bootcon_full}.}
    \label{tab:bootcon_mean_matched}
    \begin{tabular*}{\linewidth}{@{\extracolsep{\fill}}llr|rrr|r@{}}
        \toprule
        Backbone & Reference & $c_{\mathrm{match}}$ & Vanilla
        & Adaptive & BootCon & RI (\%) \\
        \midrule
        \multicolumn{7}{c}{\textbf{TUM camera pose, prefix 800; ATE $\downarrow$}} \\
        \addlinespace[0.5pt]
        CUT3R & TTT3R & 0.33 & 0.167 & 0.090 & 0.095 & 93.900 \\
        CUT3R & FILT3R & 0.15 & 0.167 & 0.056 & 0.062 & 94.449 \\
        ZipMap & FILT3R(Z) & 0.13 & 0.214 & 0.065 & 0.061 & 102.198 \\
        \midrule
        \multicolumn{7}{c}{\textbf{ScanNet camera pose, prefix 1000; ATE $\downarrow$}} \\
        \addlinespace[0.5pt]
        CUT3R & TTT3R & 0.28 & 0.834 & 0.391 & 0.414 & 94.827 \\
        CUT3R & FILT3R & 0.15 & 0.834 & 0.270 & 0.298 & 95.117 \\
        ZipMap & FILT3R(Z) & 0.13 & 1.062 & 0.494 & 0.486 & 101.325 \\
        \midrule
        \multicolumn{7}{c}{\textbf{Bonn video depth, prefix 500; Abs Rel $\downarrow$}} \\
        CUT3R & TTT3R & 0.32 & 0.086 & 0.077 & 0.077 & 101.517 \\
        CUT3R & FILT3R & 0.15 & 0.086 & 0.076 & 0.073 & 123.631 \\
        ZipMap & FILT3R(Z) & 0.13 & 0.108 & 0.063 & 0.062 & 101.172 \\
        \midrule
        \multicolumn{7}{c}{\textbf{KITTI video depth, prefix 500; Abs Rel $\downarrow$}} \\
        CUT3R & TTT3R & 0.38 & 0.130 & 0.118 & 0.118 & 99.297 \\
        CUT3R & FILT3R & 0.16 & 0.130 & 0.117 & 0.117 & 100.793 \\
        ZipMap & FILT3R(Z) & 0.14 & 0.135 & 0.088 & 0.089 & 98.674 \\
        \midrule
        \multicolumn{7}{c}{\textbf{7-Scenes 3D reconstruction, prefix 1000; Acc $\downarrow$}} \\
        CUT3R & TTT3R & 0.30 & 0.237 & 0.137 & 0.139 & 97.999 \\
        CUT3R & FILT3R & 0.15 & 0.237 & 0.058 & 0.057 & 100.328 \\
        ZipMap & FILT3R(Z) & 0.13 & 0.315 & 0.084 & 0.074 & 104.193 \\
        \midrule
        \multicolumn{7}{c}{\textbf{NRGBD 3D reconstruction, prefix 500; Acc $\downarrow$}} \\
        CUT3R & TTT3R & 0.32 & 0.308 & 0.126 & 0.129 & 98.357 \\
        CUT3R & FILT3R & 0.15 & 0.308 & 0.079 & 0.076 & 101.294 \\
        ZipMap & FILT3R(Z) & 0.13 & 0.331 & 0.041 & 0.039 & 100.669 \\
        \bottomrule
    \end{tabular*}
\end{table}

%% file: figures/stability_plasticity.tex
\begin{figure}[tp]
    \centering
    \includegraphics[width=\linewidth,page=1,trim=-4bp 0 0 0,clip]{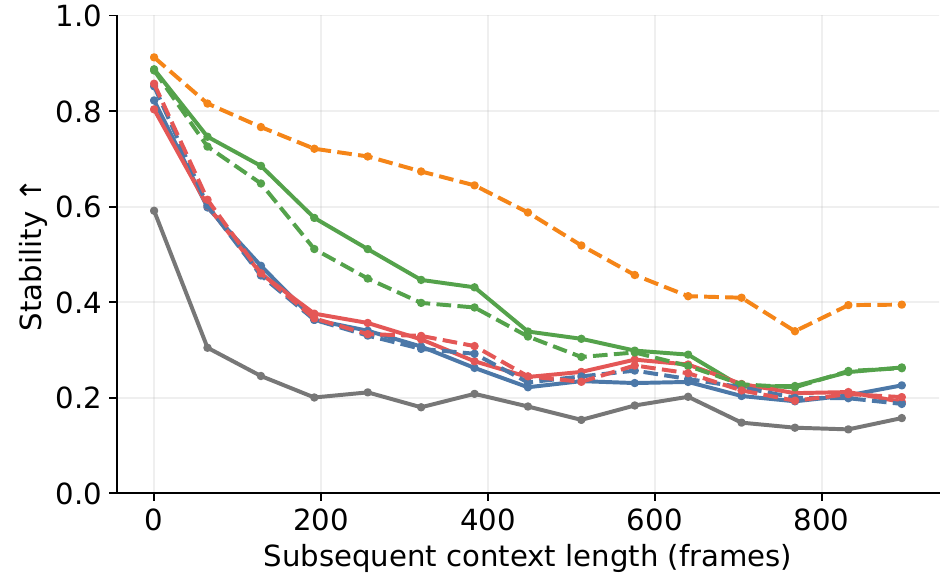}\par
    \vspace{1pt}
    \includegraphics[width=\linewidth,page=2,trim=-4bp 0 0 0,clip]{figures/stability_plasticity_artwork.pdf}\par
    \vspace{2pt}
    \begingroup
    \scriptsize\sffamily
    \setlength{\tabcolsep}{0pt}
    \renewcommand{\arraystretch}{0.95}
    \newcommand{\diagcolor}[1]{\textcolor[rgb]{#1}{\raisebox{0.45ex}{\rule{8pt}{0.7pt}}}}
    \newcommand{\diagdash}{\raisebox{0.45ex}{\rule{2.1pt}{0.7pt}\hspace{1.1pt}\rule{2.1pt}{0.7pt}\hspace{1.1pt}\rule{2.1pt}{0.7pt}}}
    \begin{tabular*}{\linewidth}{@{\extracolsep{\fill}}llll@{}}
        \diagcolor{0.4666666667,0.4666666667,0.4666666667}\,CUT3R &
        \diagcolor{0.2980392157,0.4705882353,0.6588235294}\,TTT3R &
        \diagcolor{0.3294117647,0.6352941176,0.2941176471}\,FILT3R &
        \diagcolor{0.8941176471,0.3411764706,0.337254902}\,TTSA3R
    \end{tabular*}\par
    \begin{tabular*}{\linewidth}{@{\extracolsep{\fill}}lll@{}}
        \diagcolor{0,0,0}\,Reference &
        \diagdash\,Mean-matched BootCon &
        \textcolor[rgb]{0.9607843137,0.5215686275,0.0941176471}{\diagdash}\,BootCon ($c_{\mathrm{ctx}}$)
    \end{tabular*}\par
    \endgroup
    \caption{\textbf{Stability and plasticity probing.}
    \textbf{Top}: stability measures whether later memory preserves earlier camera
    predictions.
    \textbf{Bottom}: plasticity measures the benefit of learning from new observations, i.e., capability of learning. Negative values mean writes increase error.
    Mean-matched BootCon broadly follows its adaptive references' stability-plasticity patterns.
    Our BootCon($c_{\mathrm{ctx}}$) (orange) achieves the highest
    mean stability at every tested context length and the highest mean
    plasticity at most lengths. Probe definitions and settings are in
    Appendix~\ref{app:diagnostics}.}
    \label{fig:diagnostics}
\end{figure}

%% file: figures/stability_reprediction.tex
\begin{figure}[t]
    \centering
    \includegraphics[width=\linewidth,clip]{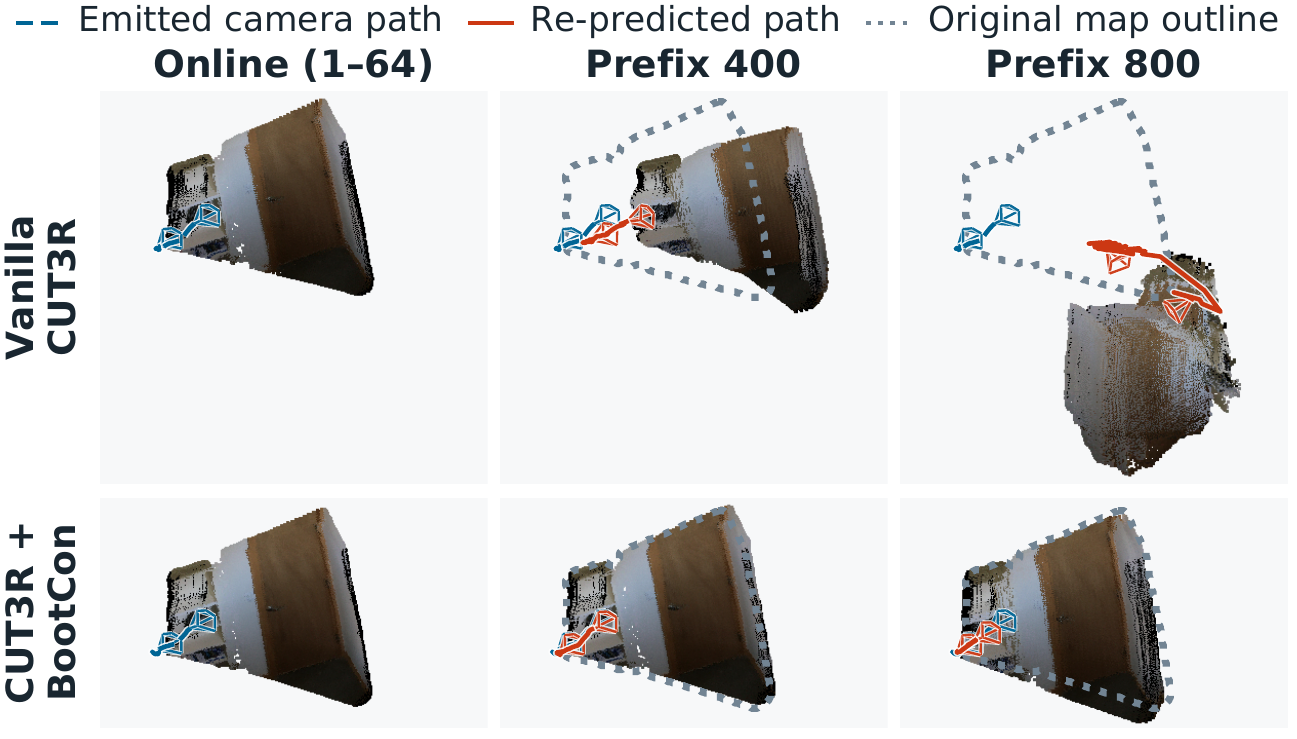}
    \caption{\textbf{Visualizing historical drift.}
The first column shows online predictions for the first 64 frames of a
ScanNet scene; later columns re-predict those same frames
after 400 and 800 total frames. Under vanilla CUT3R (top), the camera path and
doorway drift away from their original positions. BootCon
($c_{\mathrm{ctx}}$) (bottom) keeps both closer to their original positions.
Probe definitions and settings are in Appendix~\ref{app:diagnostics}.}
    \label{fig:stability_reprediction}
\end{figure}

%% file: tables/context_ratio_comparison.tex
\begin{table*}[!t]
\centering
\small
\renewcommand{\arraystretch}{1.0}
\setlength{\tabcolsep}{2.5pt}
\setlength{\aboverulesep}{0.65pt}
\setlength{\belowrulesep}{0.65pt}
\captionsetup{font=small,skip=2pt}
\caption{\textbf{Comparing BootCon($c_{\mathrm{ctx}}$) with existing methods.}
We evaluate camera-pose estimation on TUM and ScanNet, video-depth prediction
on Bonn and KITTI, and 3D reconstruction on 7-Scenes and NRGBD.
BootCon uses either a benchmark-specific design horizon or 1,000 frames on every
benchmark. With either choice, a BootCon variant outperforms all nine comparison
methods on the primary metric of each benchmark. Reconstruction entries give
mean / median. Bold marks the best value for each metric before rounding.
Evaluation settings are in Appendix~\ref{app:bootcon_context_comparison}.}
    \label{tab:bootcon_context_comparison}
\begin{tabular*}{\linewidth}{@{\extracolsep{\fill}}lrrrrrrrrr@{}}
\toprule
& \multicolumn{3}{c}{TUM pose, prefix 800}
& \multicolumn{3}{c}{Bonn depth, prefix 500}
& \multicolumn{3}{c}{7-Scenes recon., prefix 1000}\\
\cmidrule(lr){2-4}\cmidrule(lr){5-7}\cmidrule(l){8-10}
Method & ATE $\downarrow$ & RPE-T $\downarrow$ & RPE-R $\downarrow$
& Abs Rel $\downarrow$ & $\delta<1.25$ $\uparrow$ & log RMSE $\downarrow$
& Acc $\downarrow$ & Comp $\downarrow$ & NC $\uparrow$\\
\midrule
CUT3R~\citep{wang2025cut3r} & 0.167 & \textbf{0.007} & 0.489 & 0.086 & 93.638 & 0.144 & 0.237 / 0.164 & 0.102 / 0.016 & 0.512 / 0.515\\
TTT3R~\citep{chen2026ttt3r} & 0.090 & 0.010 & 0.392 & 0.077 & 95.318 & 0.137 & 0.137 / 0.084 & 0.049 / 0.011 & 0.524 / 0.533\\
FILT3R~\citep{jin2026filt3r} & 0.056 & 0.010 & 0.364 & 0.076 & 95.438 & 0.137 & 0.058 / 0.030 & 0.029 / 0.005 & 0.534 / 0.549\\
TTSA3R~\citep{zheng2026ttsa3r} & 0.077 & 0.009 & 0.369 & 0.075 & 95.439 & 0.136 & 0.095 / 0.053 & 0.037 / 0.006 & 0.529 / 0.541\\
AFG-Img~\citep{ren2026afg} & 0.068 & 0.009 & 0.359 & 0.071 & 95.584 & 0.133 & 0.048 / 0.026 & 0.025 / 0.004 & 0.535 / 0.550\\
AFG-Pose~\citep{ren2026afg} & 0.044 & 0.010 & 0.461 & 0.071 & 95.386 & 0.134 & 0.022 / 0.008 & 0.019 / 0.002 & 0.534 / 0.549\\
ReCal3R~\citep{li2026recal3r} & 0.054 & 0.009 & 0.395 & 0.072 & 95.336 & 0.135 & 0.038 / 0.018 & 0.025 / 0.004 & \textbf{0.536} / \textbf{0.552}\\
MeMix~\citep{dong2026memix} & 0.152 & 0.008 & 0.457 & 0.087 & 93.791 & 0.145 & 0.215 / 0.150 & 0.095 / 0.030 & 0.514 / 0.518\\
ZipMap~\citep{jin2026zipmap} & 0.214 & 0.017 & 9.989 & 0.108 & 93.993 & 0.166 & 0.315 / 0.167 & 0.197 / 0.135 & 0.515 / 0.520\\
\midrule
\multicolumn{10}{@{}l}{\emph{Benchmark-specific design horizon} ($c=L_{\mathrm{train}}/L_{\mathrm{infer}}$)}\\
CUT3R + BootCon & 0.048 & 0.010 & 0.585 & 0.073 & 95.366 & 0.135 & 0.021 / 0.009 & 0.021 / 0.003 & 0.536 / 0.552\\
ZipMap + BootCon & 0.037 & 0.016 & \textbf{0.350} & 0.061 & 96.501 & 0.127 & \textbf{0.017} / \textbf{0.006} & \textbf{0.019} / \textbf{0.002} & 0.532 / 0.546\\
\midrule
\multicolumn{10}{@{}l}{\emph{Shared 1,000-frame design horizon} ($c=L_{\mathrm{train}}/1000$)}\\
CUT3R + BootCon & 0.047 & 0.010 & 0.593 & 0.072 & 95.319 & 0.136 & 0.021 / 0.009 & 0.021 / 0.003 & 0.536 / 0.552\\
ZipMap + BootCon & \textbf{0.037} & 0.016 & 0.352 & \textbf{0.060} & \textbf{96.517} & \textbf{0.127} & \textbf{0.017} / \textbf{0.006} & \textbf{0.019} / \textbf{0.002} & 0.532 / 0.546\\
\bottomrule
\end{tabular*}
\par\vspace{1.5pt}
\begin{tabular*}{\linewidth}{@{\extracolsep{\fill}}lrrrrrrrrr@{}}
\toprule
& \multicolumn{3}{c}{ScanNet pose, prefix 1000}
& \multicolumn{3}{c}{KITTI depth, prefix 500}
& \multicolumn{3}{c}{NRGBD recon., prefix 500}\\
\cmidrule(lr){2-4}\cmidrule(lr){5-7}\cmidrule(l){8-10}
Method & ATE $\downarrow$ & RPE-T $\downarrow$ & RPE-R $\downarrow$
& Abs Rel $\downarrow$ & $\delta<1.25$ $\uparrow$ & log RMSE $\downarrow$
& Acc $\downarrow$ & Comp $\downarrow$ & NC $\uparrow$\\
\midrule
CUT3R & 0.834 & 0.035 & 1.249 & 0.130 & 86.581 & 0.175 & 0.308 / 0.209 & 0.175 / 0.094 & 0.551 / 0.577\\
TTT3R & 0.391 & 0.039 & 1.279 & 0.118 & 89.265 & 0.164 & 0.126 / 0.063 & 0.031 / 0.003 & 0.584 / 0.630\\
FILT3R & 0.270 & 0.042 & 1.360 & 0.117 & 89.562 & 0.163 & 0.079 / 0.030 & 0.018 / 0.003 & 0.597 / 0.652\\
TTSA3R & 0.378 & 0.038 & 1.254 & 0.117 & 89.546 & 0.163 & 0.085 / 0.034 & 0.020 / 0.003 & 0.593 / 0.645\\
AFG-Img & 0.285 & 0.046 & 1.869 & 0.117 & 89.504 & 0.162 & 0.066 / 0.025 & 0.011 / 0.003 & 0.597 / 0.653\\
AFG-Pose & 0.285 & 0.059 & 2.620 & 0.118 & 89.349 & 0.163 & 0.057 / 0.020 & 0.010 / 0.002 & 0.590 / 0.638\\
ReCal3R & 0.267 & 0.047 & 2.050 & 0.117 & 89.613 & 0.162 & 0.069 / 0.025 & 0.013 / 0.003 & 0.596 / 0.652\\
MeMix & 0.764 & \textbf{0.033} & \textbf{1.048} & 0.127 & 86.962 & 0.173 & 0.259 / 0.160 & 0.183 / 0.142 & 0.560 / 0.591\\
ZipMap & 1.062 & 0.095 & 9.476 & 0.135 & 83.384 & 0.187 & 0.331 / 0.189 & 0.126 / 0.077 & 0.563 / 0.595\\
\midrule
\multicolumn{10}{@{}l}{\emph{Benchmark-specific design horizon} ($c=L_{\mathrm{train}}/L_{\mathrm{infer}}$)}\\
CUT3R + BootCon & 0.263 & 0.049 & 2.256 & 0.117 & 89.365 & 0.163 & 0.072 / 0.027 & 0.015 / 0.003 & \textbf{0.598} / \textbf{0.655}\\
ZipMap + BootCon & \textbf{0.250} & 0.069 & 2.112 & 0.090 & 93.055 & 0.138 & 0.026 / 0.014 & \textbf{0.007} / 0.002 & 0.579 / 0.619\\
\midrule
\multicolumn{10}{@{}l}{\emph{Shared 1,000-frame design horizon} ($c=L_{\mathrm{train}}/1000$)}\\
CUT3R + BootCon & 0.263 & 0.056 & 2.659 & 0.118 & 89.446 & 0.163 & 0.060 / 0.021 & 0.010 / 0.003 & 0.597 / 0.653\\
ZipMap + BootCon & 0.271 & 0.081 & 3.106 & \textbf{0.089} & \textbf{93.859} & \textbf{0.137} & \textbf{0.025} / \textbf{0.014} & 0.007 / \textbf{0.002} & 0.577 / 0.615\\
\bottomrule
\end{tabular*}
\end{table*}

%% file: figures/camera_trajectories.tex
\begin{figure*}[tp]
    \centering
    \includegraphics[width=\linewidth,height=0.78\textheight,keepaspectratio]{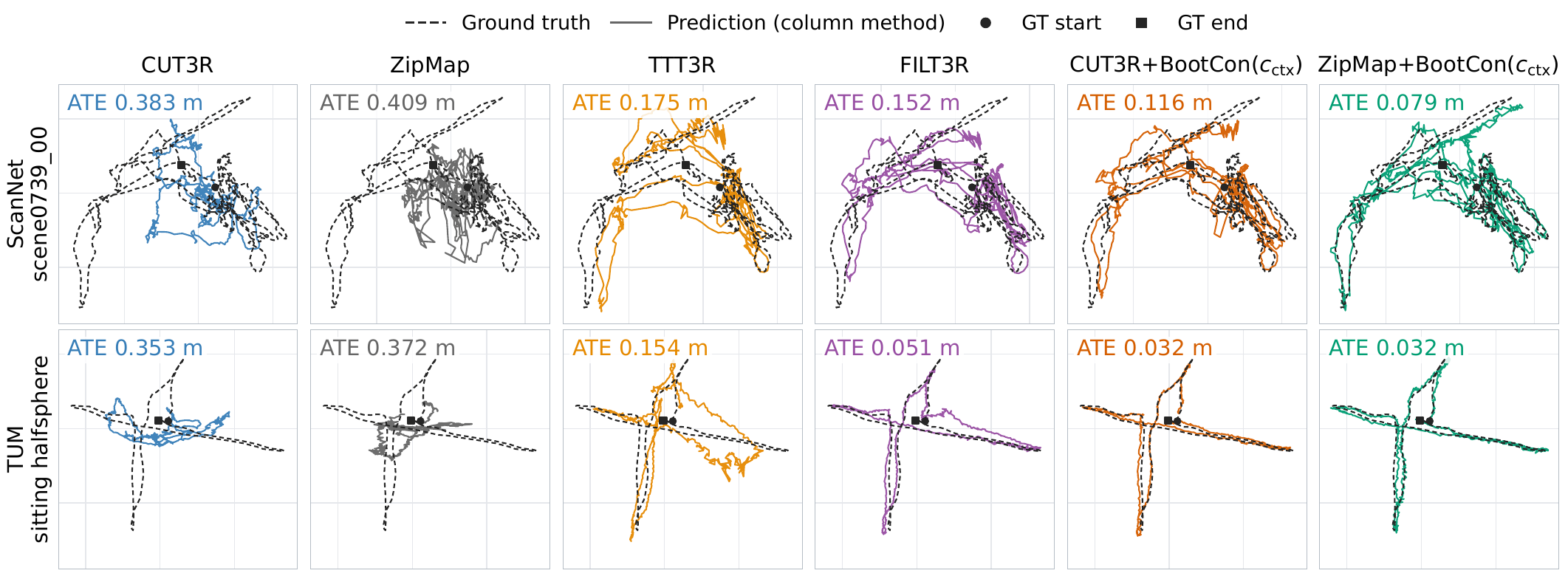}
    \caption{\textbf{Qualitative Camera Trajectory Results.}
We show a ScanNet trajectory (top) and a TUM trajectory(bottom). 
In both examples, BootCon($c_\mathrm{ctx}$) on either backbone follows ground truth
more closely and achieves lower ATE than vanilla CUT3R, streaming ZipMap,
TTT3R, and FILT3R.
Trajectories are independently aligned to ground truth by Sim(3) and
projected onto a shared plane within each row.
More qualitative trajectories are in
Figure~\ref{fig:additional_pose_qualitative} of Supplementary Material.}
    \label{fig:scannet_pose_qualitative}
\end{figure*}

%% file: figures/reconstruction_qualitative.tex
\begin{figure*}[t]
    \centering
    \includegraphics[width=\linewidth,clip]{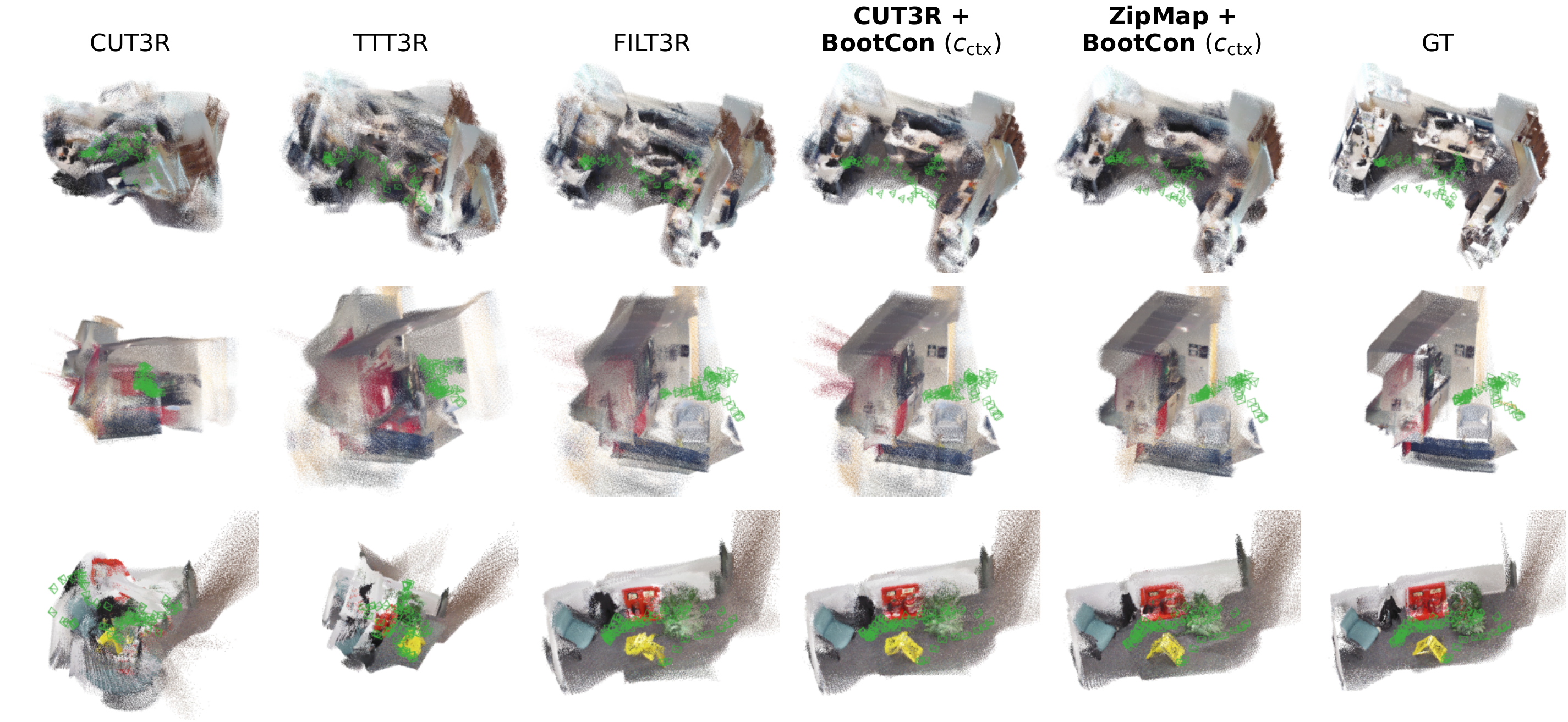}
    \caption{\textbf{Qualitative 3D Reconstruction Results.}
We show three sample reconstructed point maps, together with the ground truths. BootCon models based on both backbones reduce the distortions
visible under CUT3R and produces geometry visually significantly better than adaptive methods, such as TTT3R and FILT3R. More qualitative results are in Supplementary Material.}
    \label{fig:reconstruction_qualitative}
\end{figure*}

%% file: tables/bootstrap_ablation.tex
\begin{table}[tbp]
    \centering
    \fontsize{8}{9}\selectfont
    \renewcommand{\arraystretch}{1.05}
    \setlength{\tabcolsep}{2.5pt}
    \setlength{\aboverulesep}{1pt}
    \setlength{\belowrulesep}{1pt}
    \captionsetup{font=small,skip=3pt}
    \caption{\textbf{Ablations on BootCon($c_{\mathrm{ctx}}$).} Using ZipMap-base BootCon($c_{\mathrm{ctx}}$) on TUM, we conduct:
    (a) \emph{Bootstrap Ablations}. While keeping the later update weights as 
    $c_{\mathrm{ctx}}=0.03$, omitting the full bootstrap or tripling its
    weight worsens ATE, while two or three initial full updates yield modest
    gains. (b) \emph{Conservative Update Ablations}. All variants retain a
    full bootstrap and a post-bootstrap mean of $0.03$. Periodic and randomly
    placed full updates are slightly worse than BootCon, whereas uniform
    random framewise weights, rescaled to the same mean, perform similarly.
    Random variants report mean $\pm$ std over five seeds.}
    \label{tab:bootcon_bootstrap_ablation}
    \begin{minipage}[t]{0.52\linewidth}
    \vspace{0pt}
    \centering\textbf{(a) Bootstrap}\par\vspace{2pt}
    \begin{tabular*}{\linewidth}{@{\extracolsep{\fill}}lr@{}}
        \toprule
        Variant & ATE (m) $\downarrow$ \\
        \midrule
        BootCon($c_{\mathrm{ctx}}$) & 0.0372 \\
        \midrule
        No bootstrap ($c_1=c_{\mathrm{ctx}}$) & 0.0552 \\
        Stronger ($c_1=2$) & 0.0374 \\
        Stronger ($c_1=3$) & 0.0537 \\
        Two-frame ($c_{1:2}=1$) & 0.0354 \\
        Three-frame ($c_{1:3}=1$) & 0.0356 \\
        \bottomrule
    \end{tabular*}
    \end{minipage}\hfill
    \begin{minipage}[t]{0.47\linewidth}
    \vspace{0pt}
    \centering\textbf{(b) Conservative Update}\par\vspace{2pt}
    \begin{tabular*}{\linewidth}{@{\extracolsep{\fill}}lr@{}}
        \toprule
        Variant & ATE (m) $\downarrow$ \\
        \midrule
        BootCon($c_{\mathrm{ctx}}$) & 0.0372 \\
        \midrule
        Periodic full & 0.0384 \\
        Random full & $0.0383\pm0.0018$ \\
        Random $c_t$ & $0.0371\pm0.0002$ \\
        \bottomrule
    \end{tabular*}
    \end{minipage}
\end{table}

%% file: tables/mean_matched_relative_pose.tex
\begin{table}[!ht]
\centering\footnotesize
\setlength{\tabcolsep}{2.5pt}
\caption{\textbf{Local camera motion after removing adaptive modulation.}
Mean-matched BootCon preserves or improves
local motion accuracy in three of four method--dataset pairs. Only FILT3R
on ScanNet benefits from adaptive modulation on both metrics.
We report one-frame translation (T, m) and rotation (R, $^\circ$) errors
for the CUT3R-based methods at the prefixes used in Table~\ref{tab:bootcon_mean_matched}.
A dash means RI is inapplicable because the adaptive error equals or exceeds
the vanilla error. RI uses unrounded errors.}
\label{tab:bootcon_rpe_matched}
\begin{tabular}{@{}llrrrr@{}}
\toprule
Reference & & Vanilla & Adaptive & BootCon & RI (\%)\\
\midrule
\multicolumn{6}{c}{\textbf{TUM camera pose, prefix 800}} \\
TTT3R & T & 0.007 & 0.010 & 0.009 & ---\\
($c=0.33$) & R & 0.489 & 0.392 & 0.388 & 103.838 \\
FILT3R & T & 0.007 & 0.010 & 0.009 & ---\\
($c=0.15$) & R & 0.489 & 0.364 & 0.357 & 105.796 \\
\midrule
\multicolumn{6}{c}{\textbf{ScanNet camera pose, prefix 1000}} \\
TTT3R & T & 0.035 & 0.039 & 0.039 & ---\\
($c=0.28$) & R & 1.249 & 1.279 & 1.239 & --- \\
FILT3R & T & 0.035 & 0.042 & 0.044 & ---\\
($c=0.15$) & R & 1.249 & 1.360 & 1.806 & --- \\
\bottomrule
\end{tabular}
\end{table}

%% file: figures/camera_trajectories_additional.tex
\begin{figure*}[tp]
    \centering
    \includegraphics[width=\linewidth,height=0.78\textheight,keepaspectratio]{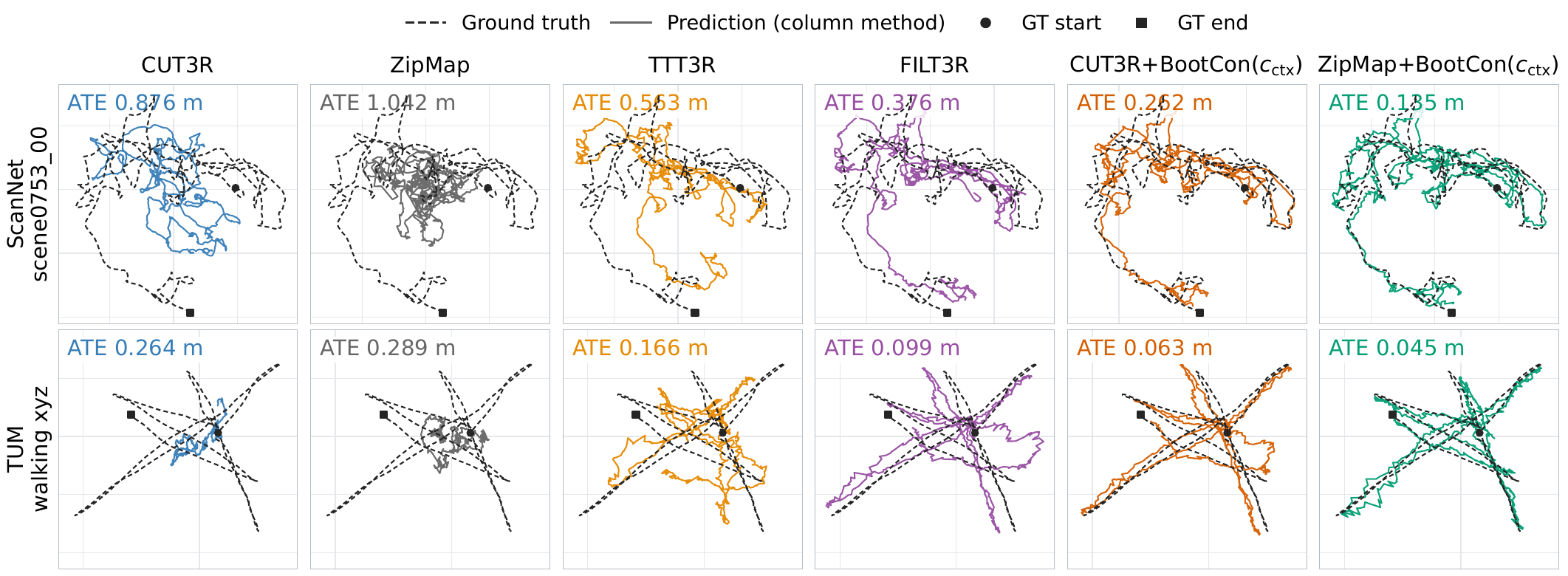}
    \caption{\textbf{Additional Qualitative Camera Trajectory Results.}
    The top row shows ScanNet \texttt{scene0753\_00} at 1,000 frames
    (stride 3), and the bottom row shows TUM \texttt{walking\_xyz}
    at 800 frames (stride 1).
    In both scenes, BootCon on either backbone has lower ATE than the four
    comparison methods, with ZipMap-based BootCon closest to ground truth.
    Update weights, alignment, projection, and visualization conventions
    follow Figure~\ref{fig:scannet_pose_qualitative}. Both axes are in meters,
    and ATE is measured in 3D.}
    \label{fig:additional_pose_qualitative}
\end{figure*}

%% file: figures/context_ratio_sweep.tex
\begin{figure*}[tp]
    \centering
    \includegraphics[width=\linewidth]{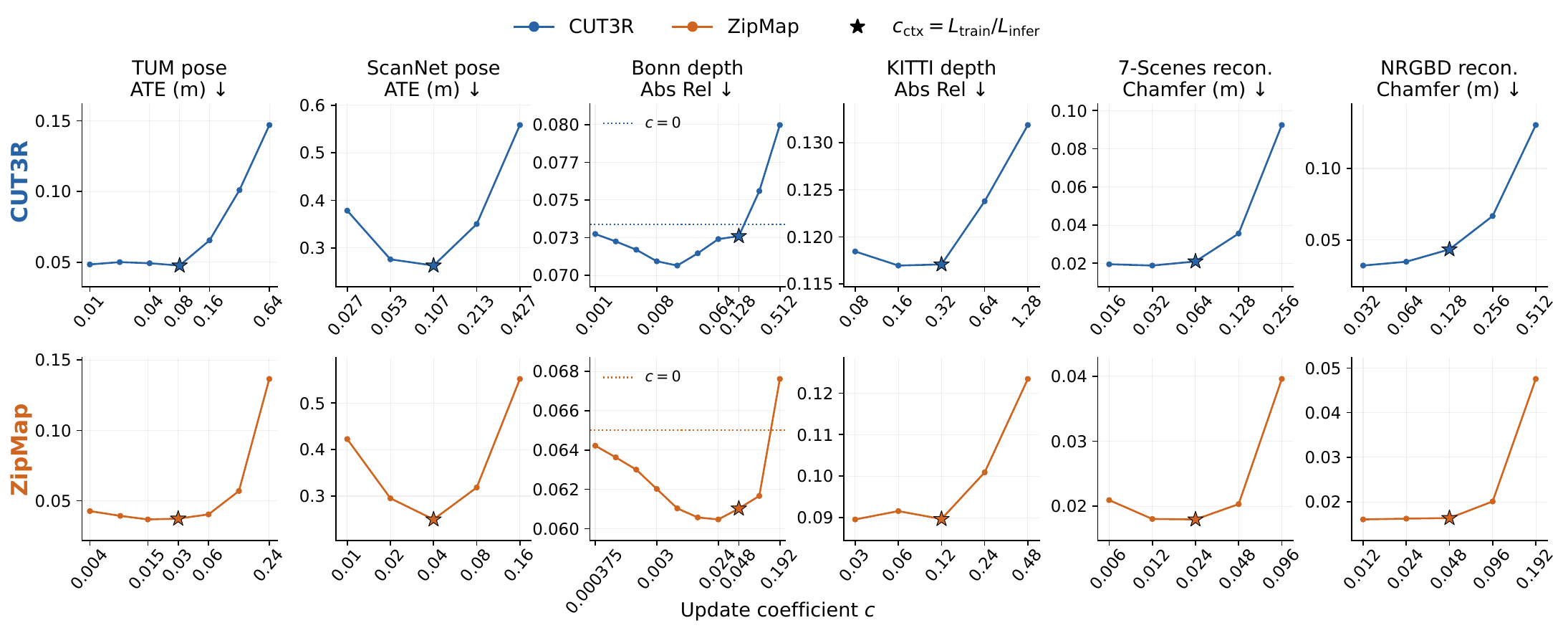}
    \caption{\textbf{$c$ sweeping around $c_{\mathrm{ctx}}$.}
    Top: CUT3R; bottom: ZipMap. Stars mark $c_{\mathrm{ctx}}$;
    lower error is better. The coefficient axis is logarithmic.}
    \label{fig:bootcon_selection}
\end{figure*}

%% file: tables/context_ratio_settings.tex
\begin{table}[htbp]
\centering\small
\caption{\textbf{Benchmarks and Their Settings for Table~\ref{tab:bootcon_context_comparison}.}
For each benchmark, we report the following settings.
(1)~$N$ is the number of evaluated sequences.
(2)~\emph{Cap} is the maximum number of processed frames scored per sequence.
(3)~\emph{Median length} is the median sequence length in processed frames
after applying the cap.
(4)~$L_{\mathrm{infer}}$ is the benchmark-specific design horizon, equal to that
median rounded up to a multiple of 100 frames.
(5)~\emph{CUT3R $c_{\mathrm{ctx}}$} and \emph{ZipMap $c_{\mathrm{ctx}}$} are BootCon's constant update
weights after the first frame, $64/L_{\mathrm{infer}}$ and
$24/L_{\mathrm{infer}}$, respectively.}
\label{tab:context_comparison_settings}
\setlength{\tabcolsep}{2pt}
\begin{tabular}{@{}lrrrrrr@{}}
\toprule
Benchmark & $L_{\mathrm{infer}}$ & Cap & \shortstack[r]{Median\\length} & $N$ &
\shortstack{CUT3R\\$c_{\mathrm{ctx}}$} & \shortstack{ZipMap\\$c_{\mathrm{ctx}}$}\\
\midrule
TUM pose & 800 & 800 & 800 & 8 & 0.080 & 0.030\\
ScanNet pose & 600 & 1000 & 556 & 67 & $64/600$ & 0.040\\
Bonn depth & 500 & 500 & 500 & 5 & 0.128 & 0.048\\
KITTI depth & 200 & 500 & 144 & 13 & 0.320 & 0.120\\
7-Scenes recon. & 1000 & 1000 & 1000 & 18 & 0.064 & 0.024\\
NRGBD recon. & 500 & 500 & 500 & 9 & 0.128 & 0.048\\
\bottomrule
\end{tabular}
\end{table}

%% file: figures/prefix_benchmarks.tex
\begin{figure*}[t]
    \centering
    \includegraphics[width=\linewidth]{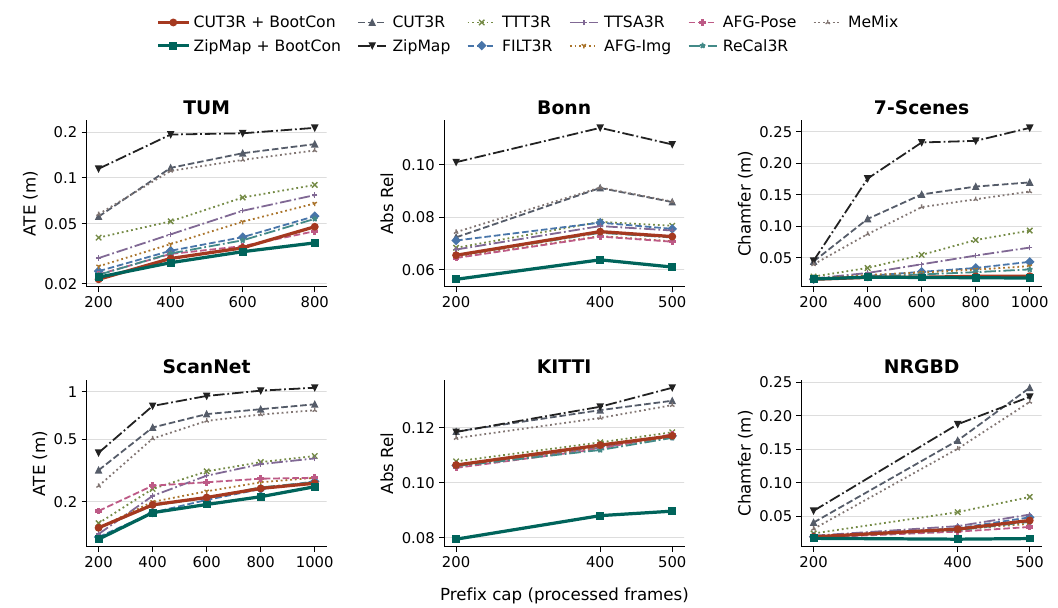}
    \caption{\textbf{Performance across prefix lengths.}
    We compare the nine baselines and the two benchmark-specific BootCon rows
    from Table~\ref{tab:bootcon_context_comparison}.
    From left to right, we report pose ATE (log scale), depth Abs Rel, and
    reconstruction Chamfer distance (the mean of accuracy and completeness).
    Lower is better in every panel.
    Every point on a given BootCon curve uses the same post-bootstrap weight
    $c_{\mathrm{ctx}}$, computed once for that backbone and benchmark.
    Both BootCon variants improve on their respective backbones at every measured prefix.
    Across the eleven methods, a BootCon variant has the lowest error
    at all six benchmark endpoints and in 16 of the 17 intermediate comparisons.
    The exception is 7-Scenes at 200 frames, where ReCal3R narrowly leads.}
    \label{fig:prefix_benchmarks}
\end{figure*}